\documentclass[journal=jacsat,manuscript=article]{achemso}

\usepackage[version=3]{mhchem} 
\usepackage{amssymb}

\author{Peter Pak}
\affiliation{
  Department of Mechanical Engineering, Carnegie Mellon University, Pittsburgh,
  PA, USA
}
\author{Amir Barati Farimani}
\email{barati@cmu.edu}
\affiliation{
  Department of Mechanical Engineering, Carnegie Mellon University, Pittsburgh,
  PA, USA
}
\alsoaffiliation{
  Machine Learning Department, Carnegie Mellon University, Pittsburgh, PA, USA
}

\title[]{Domain Adapted Large Language Models for Additive Manufacturing}

\abbreviations{IR,NMR,UV}

\SectionNumbersOn
\begin{document}

\begin{abstract}
This work presents a collection of multi-modal domain adapted large language
models built upon the instruction tuned variants of open weight models (Gemma 3,
Qwen 3, Gemma 4) using a relatively small dataset of around 50 million tokens.
The dataset consists of open-access additive manufacturing journal articles with
data extracted for the domain adaptive pretraining and visual instruction tuning
processes. Various stages of the developed model are evaluated with the
\texttt{Additive-\allowbreak Manufacturing-\allowbreak Benchmark} which consists
of additive manufacturing domain specific tasks compiled published resources.
Domain adapted and instruction tuned models exhibit proficiency in both language
and vision based tasks, achieving accuracies upwards of 90\% in general additive
manufacturing knowledge. This domain adaptive pretraining and instruction tuning
strategy outline an accessible specialization method for large language models
to a domain such as additive manufacturing.
\end{abstract}


\section{Introduction}
Large language Models (LLMs) have exhibited proficient capability for knowledge
based tasks in fields extending beyond that of natural language processing such
as chemistry \cite{zeng_llm-guided_2025, ock_adsorb-agent_2026,
balaji_gpt-molberta_2023}, design \cite{jadhav_large_2026, wang_human-llm_2025,
zhou_exploring_2024, xu_llm_2024}, mathematics \cite{pei_mathfusion_2025,
shojaee_llm-sr_2024, lorsung_explain_2025, kumarappan_leanagent_2025}, robotics
\cite{george_llm_2025, arnab_vivit_2021, barkley_semantic_2025,
barkley_synthesizing_2026, bartsch_llm-craft_2025}, and software
development\cite{he_llm-based_2025, xia_agentless_2024, han_tdflow_2026}. Within
Additive Manufacturing (AM), LLMs have been applied to various tasks such as
knowledge retrieval \cite{chandrasekhar_amgpt_2024,
naghavi_khanghah_multimodal_2025}, parameter selection
\cite{pak_additivellm_2025, pak_agentic_2026}, and process optimization
\cite{jadhav_llm-3d_2025}. When applied to agentic systems, the reasoning and
strategic planning capabilities of LLMs are particularly useful for enabling the
intelligent automation of complex tasks such as drug discovery
\cite{ock_large_2025}, alloy evaluation \cite{pak_agentic_2026}, print
optimization \cite{jadhav_llm-3d_2025}, and code debugging
\cite{han_tdflow_2026} to name a few. To more efficiently execute these
specialized tasks, an LLM aware of the complexities and discoveries within the
selected field is desired\cite{gururangan_dont_2020, bajan_exploring_2025,
lu_fine-tuning_2025}.

To this point, domain adapted LLMs have a number of unique advantages over
generally purpose LLMs namely the efficiency in which valid responses can be
generated by drawing on parametric (within training set)
data\cite{gururangan_dont_2020, lu_fine-tuning_2025}. Non-parametric data can be
injected into the LLM's response generation process through the use Retrieval
Augmented Generation (RAG) based architectures
\cite{lewis_retrieval-augmented_2020}. This is a powerful approach that is
capable of generating results grounded in factual data
\cite{chandrasekhar_amgpt_2024, naghavi_khanghah_multimodal_2025}, however, with
each request the retrieved passages consume additional context within the
conversation window. For dynamic data (events, machine logs, database updates,
etc.) RAG enables up-to-date responses without additional pretraining
\cite{lewis_retrieval-augmented_2020}. The medium for domain knowledge is
generally static (journal articles, figures, textbooks, etc.) and the extra
training cost done with Domain Adaptive Pretraining (DAPT) enables accurate
response generation without additional context consumption seen with
RAG\cite{gururangan_dont_2020}. Another consideration is the local use and
deployment of these LLMs as seen in cases where national security
\cite{chen_survey_2025, caballero_large_2024}, patient data
\cite{montagna_privacy-preserving_2025, qu_novel_2024, tahera_sok_2026}, or
environmental limitations \cite{barkley_synthesizing_2026} deem an edge or
on-premise solution necessary.

\begin{figure}
    \centering
    \includegraphics[width=\textwidth]{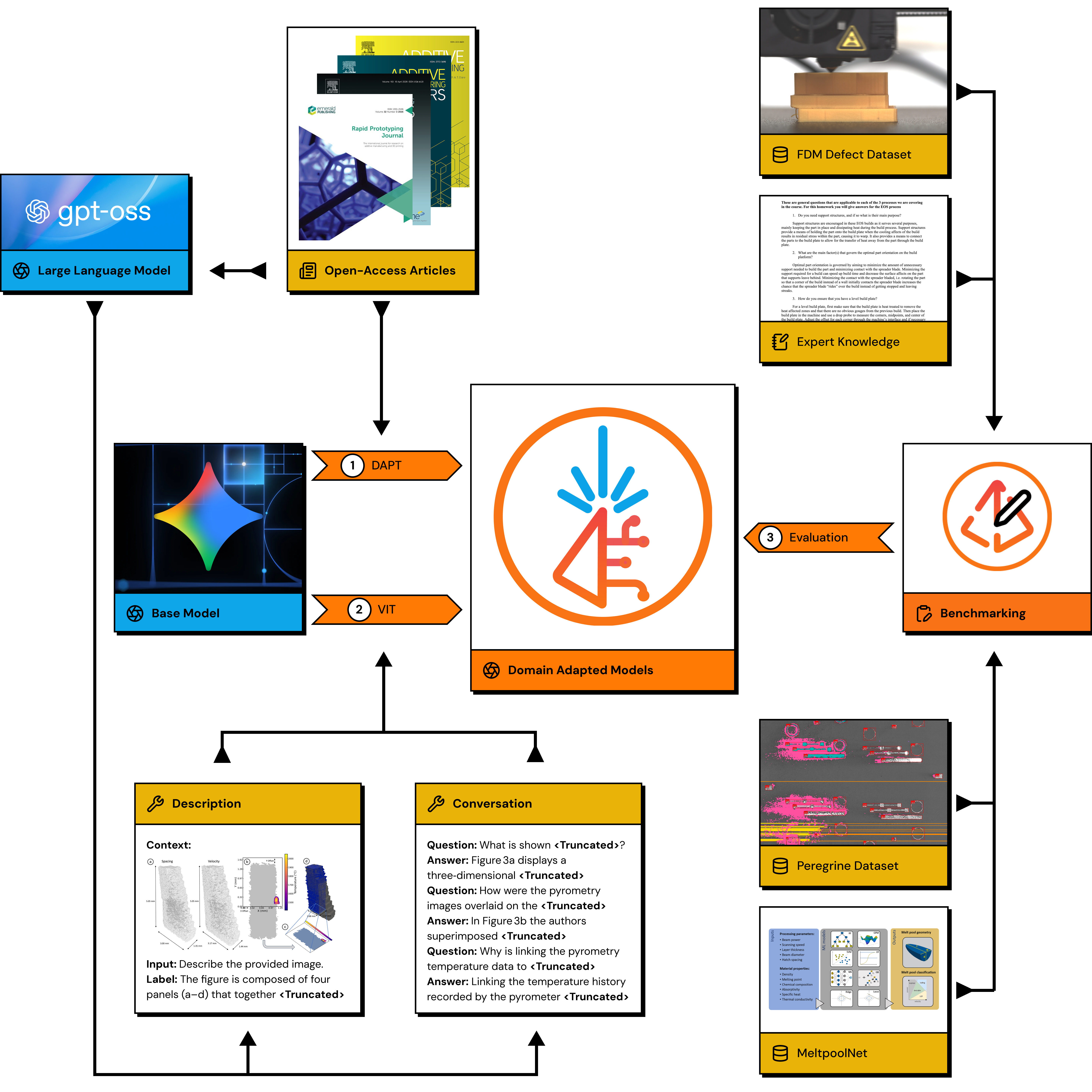}
    \caption{
        General training and evaluation process utilized in the development of
        various additive manufacturing specific large language models.
        \textbf{(1)} Domain adaptive pretraining (DAPT) is applied in both the
        text and vision modalities using the Gemma 3 model using a selection of
        open-access articles from the additive manufacturing domain.
        \cite{team_gemma_2025} as its base. \textbf{(2)} Visual Instruction
        Tuning (VIT) utilizes the extracted captions from these articles and
        generates description and conversation prompts with the aid of GPT-OSS
        120B \cite{openai_gpt-oss-120b_2025}. The domain adapted model is
        evaluated with tasks from the \texttt{Additive-\allowbreak
        Manufacturing-\allowbreak Benchmark} which are compiled from various
        published resources \cite{akbari_meltpoolnet_2022,
        scime_layer-wise_2023, hu_real-time_2024}.
    }
    \label{fig:main}
\end{figure}

In this work, a set of domain adapted LLMs for additive manufacturing are
created with a collection of multi-modal open weight models including the Gemma
3 (12B) model \cite{team_gemma_2025}, Qwen 3 (8B) model \cite{yang_qwen3_2025},
and Gemma 4 (31B) model (Fig. \ref{fig:main}). Domain adaptive pretraining and
visual instruction tuning is performed using text and image data from various
open-access additive manufacturing journal articles curated within a public
dataset \cite{peter_pak_additivellm2-oa_nodate}. This work also introduces the
\texttt{Additive-\allowbreak Manufacturing-\allowbreak Benchmark} which measures
the LLM's specific capabilities in tasks such as melt pool dimensional
prediction, anomaly identification, and general knowledge about AM. The methods
and architecture used in developing these models showcase how large language
models can be efficiently tailored to provide enhanced domain knowledge within a
given field.

\section{Related Work}
Previous works have explored the application of large language models to solving
challenges within the domain of additive manufacturing. These include approaches
such as in-context learning \cite{fang_large_2024},
fine-tuning\cite{pak_additivellm_2025}, retrieval augmented generation
\cite{chandrasekhar_amgpt_2024, naghavi_khanghah_multimodal_2025}, the use of
vision language models \cite{zheng_qa-vlm_2026, liu_visual_2023,
pak_thermopore_2024}, and the development of agentic
systems\cite{jadhav_llm-3d_2025, pak_agentic_2026}. To the best of the authors'
knowledge there exists no work which investigates the adaptation of large
language models to domain of additive manufacturing within both the language and
vision space through the use of continual pre-training and instruction tuning.

AdditiveLLM \cite{pak_additivellm_2025} investigated the fine-tuning of various
different pretrained large language models and evaluated these models on their
prediction accuracy regarding the classification task of process map defect
regimes. Data used for this fine-tuning process was obtained from the
MeltpoolNet dataset \cite{akbari_meltpoolnet_2022} and FLOW-3D based
simulations. This dataset was used to fine-tune models ranging in size from 60
million parameters to 1 billion parameters: DistilBERT
(66M)\cite{sanh_distilbert_2020}, SciBERT (110M) \cite{beltagy_scibert_2019},
T5-Small (60M)\cite{raffel_exploring_2023}, and Llama 3.2
(1B)\cite{grattafiori_llama_2024}. An accuracy of 82\% was achieved when
predicting defect regimes within laser powder bed fusion (keyholing, lack of
fusion, balling, or none) given natural language formatted process parameters
using the fine-tuned DistilBERT model \cite{pak_additivellm_2025}.

An in-context learning approach to enable large language models to detect
defects encountered during the vat photopolymerization process was explored by
Fang et al. \cite{fang_large_2024} The authors utilize a camera mounted to the
underside of the resin vat to obtain images of the resin deposition process,
looking for defects such as debris and unfilled streaks within the build
platform\cite{fang_large_2024}. The taken images are then provided to GPT-4o,
along with positive and negative image samples, in order to predict whether the
current layer is normal or defective \cite{fang_large_2024}. These samples,
along with text descriptions of both cases, provides additional contextual
information to guide the large language model to predict the correct outcome
only through information provided within the conversation's context window. By
coupling the language and image descriptions of each case, this in-context
learning method can distinguish between normal and defective build layers and
achieved a 96\% classification accuracy\cite{fang_large_2024}.

In AMGPT by Chandrasekhar et al. \cite{chandrasekhar_amgpt_2024} explores the
use of Retrieval Augmented Generation (RAG)
\cite{lewis_retrieval-augmented_2020} specifically within the application of
additive manufacturing. RAG at its core is based on the cosine similarity
between the query and document passages, often utilizing separate encoders for
the document passages and query text to compare the two within the same
embedding space \cite{lewis_retrieval-augmented_2020}. The authors utilize this
dual-encoder retrieval mechanism to obtain the relevant passages for a given
specific user query and these passages are provided along with the query as
input to the pretrained Llama2-7B \cite{touvron_llama_2023} model for response
generation. This work showcases the effectiveness of utilizing empirical data
within the generation process of a large language model as the authors make the
claim that their RAG enabled response displayed less factual errors than that of
GPT-4 \cite{chandrasekhar_amgpt_2024}. Similar to this, work by Naghavi Khanghah
et al. \cite{naghavi_khanghah_multimodal_2025} extends the use of RAG into
vision space and leverages a multimodal approach for the detection and
classification of anomalies specifically within laser powder bed fusion. With
over 50 test images, measured anomalies of recoater hopping, recoater streaking,
incomplete spreading, swelling, debris, and soot were classified. Through the
use of models such as Qwen2-VL-2B and GPT-4o-mini, the authors demonstrate that
the utilization of the RAG based system was reported to improve the accuracy of
model prediction by around 12\% \cite{naghavi_khanghah_multimodal_2025}. The
domain specific capabilities of these large language models can then be utilized
in agentic systems as their multi-modal and reasoning abilities are quite
suitable for the orchestration of tool calls and actions. This approach can be
utilized in tasks such as alloy evaluation for additive manufacturing where
based on material properties calculated by Thermo-Calc, the potential for lack
of fusion porosity can be evaluated for a composition of elements
\cite{pak_agentic_2026}. Another application of agents is shown in LLM-3D Print
by Jadhav et al. \cite{jadhav_llm-3d_2025} which explores the use of a
multi-agent system for the detection and mitigation of defects within the Fused
Deposition Modeling (FDM) process. These prints were evaluated with compression
testing and showed that the agentic system helped enable clear improvements in
mechanical performance \cite{jadhav_llm-3d_2025}. In both of these works, the
agentic system relied on off-the-shelf large language models (i.e. GPT-4o,
Claude, Gemini) \cite{jadhav_llm-3d_2025, pak_agentic_2026}. However, it has yet
to be explored that a domain adapted, fine-tuned model for additive
manufacturing within these agentic systems would exhibit enhanced performance. 

\section{Background}
\subsection{Large Language Models}

A Large Language Model (LLM) is a neural network which commonly utilizes
transformer based architectures trained to the task of next token prediction
\cite{vaswani_attention_2023, devlin_bert_2019}. This type of model is often
pretrained on a corpus of natural language data ranging from Wikipedia articles
to code available on GitHub and showcases its comprehension of these datasets
through various benchmarking tasks.\cite{gururangan_dont_2020,
hendrycks_measuring_2020, cobbe_training_2021}. Adhering to scaling laws, these
models often exhibit improved performance with larger parameter counts, compute
times, and dataset size \cite{brown_language_2020}. Furthermore, these models
can operate beyond the bounds of natural language through different
architectural modifications which allows for the interpretation of images
\cite{dosovitskiy_image_2020}, videos \cite{arnab_vivit_2021}, and 3d models
\cite{yu_point-bert_2022}.

\subsubsection{Transformer Architecture}
\label{sec:transformers}
Before the inception of the transformer architecture
\cite{vaswani_attention_2023}, long short-term memory
\cite{hochreiter_long_1997} and gated recurrent neural networks
\cite{chung_empirical_2014} were the predominant approaches to solve language
and sequence modeling tasks. However, these previous approaches were limited in
their lack of parallelization and constrained context window as they struggled
to model long range dependencies \cite{hochreiter_long_1997,
chung_empirical_2014, vaswani_attention_2023}. The transformer architecture, in
contrast, primarily relying upon the attention mechanism to model long range
dependencies bypassing the need for convolution or recurrent mechanisms
\cite{vaswani_attention_2023}. Within the original transformer model, the
architecture consists of two stacks: the encoder stack and the decoder stack
\cite{vaswani_attention_2023}. The encoder stack consists of bidirectional self
attention where a contextual representation can be generated by attending to
entirely of input tokens \cite{vaswani_attention_2023}. The decoder is concerned
with next token generation as it can only attend to the previous tokens within
the output sequence \cite{vaswani_attention_2023}.

Implementations of this encoder-decoder transformer architecture include models
such as T5 \cite{raffel_exploring_2023} and BART \cite{lewis_bart_2019}, which
excelled in fixed output tasks such as summarization and translation. Encoder
only models such as BERT \cite{devlin_bert_2019} and RoBERTa
\cite{liu_roberta_2019} utilize leverage bidirectional attention to generate a
comprehensive embedding space for dense retrieval, particularly useful in domain
specific representation environments such as those covered in CatBERTa
\cite{ock_catalyst_2023}, SciBERT \cite{beltagy_scibert_2019}, and
\cite{yu_point-bert_2022}. However, decoder only transformer stacks such as GPT
\cite{radford_improving_2018} have evolved to become the dominant approach as it
scales better for generative and reasoning tasks focused next token prediction.

\subsubsection{Multi-modal Input Representation}
\label{sec:multi-modal}
The transformer architecture can be applied to multi-modal tasks through
modifications within the tokenization process allowing
for an effective representation of visual images \cite{dosovitskiy_image_2020,
arnab_vivit_2021} and 3D models \cite{yu_point-bert_2022}. Text is provided to
the transformer as 1 dimensional input vectors and naively flattening 2 or 3
dimensional data often produces inadequate representations, often resulting in
lost temporal and spatial data \cite{dosovitskiy_image_2020, arnab_vivit_2021}.
An approach to preserving spatial data is outlined in work by Dosovitskiy et al.
\cite{dosovitskiy_image_2020} where the authors split an input image into fixed
patches while applying positional embedding in their Vision Transformer (ViT).
This is further expanded upon by Arnab et al. \cite{arnab_vivit_2021} where
these patches are expanded in an additional dimension for video frames to embed
spatial and temporal information into the input \cite{arnab_vivit_2021}.

For 3D models, point cloud representation is an efficient means of representing
spatial information without the rigid constraints of voxelization. However, the
unstructured format of point clouds presents a challenge when adapting this data
to be suitable for transformer input as the tokenization process for such a
representation is not immediately obvious\cite{yu_point-bert_2022}. Point-BERT
\cite{yu_point-bert_2022} resolves this issue by partitioning the entire 3D
model into point based patches similar to the previously methologies of ViT
\cite{dosovitskiy_image_2020}. These patches of points referred to as
``sub-clouds" preserves the spatial information necessary for adequately mapping
3D model data in an format comprehensible by the transformer architecture
\cite{yu_point-bert_2022}.

\subsubsection{Model Scaling}
\label{sec:transformer-scaling}
With the increasing parameter size of models developed with the transformer
architecture, emergent behaviors such as reasoning become evident
\cite{wei_finetuned_2021, kaplan_scaling_2020, lu_learn_2022}. This evolution in
scale can be attributed to a number of factors such as the parallelization of
self-attention computations, regularization of model weights through residual
connections, and the shift to decoder focused transformer stacks
\cite{radford_improving_2018, vaswani_attention_2023}. This finding is validated
by Kaplan et al. \cite{kaplan_scaling_2020} where model performance depends
heavily upon number of parameters, size of dataset, and the amount of compute
used during training. This work established the existence of a power law
relationship between performance and factors such as parameter size (768 to
1.5B), dataset size (22M - 23B), and compute ($10^{-5}$ to $10^{3}$ PetaFlop
days) \cite{kaplan_scaling_2020}.

\subsection{Prompting and Reasoning}
\subsubsection{Chain-of-Thought}
\label{sec:cot}
Chain-of-Thought (CoT) is multi-step prompting technique to elicit further
developed answers from the large language model than simple standard
prompting\cite{kojima_large_2023, wei_chain--thought_2023}. In this method, the
prompt is formatted in a manner such that a step-by-step answer is provided to
an example question before a similar question is posed in the
input\cite{kojima_large_2023, wei_chain--thought_2023}. This facilitates
reasoning within the model as it decomposes the prompt into a multi-step problem
which allows for additional computations be allocated to these individual
steps\cite{wei_chain--thought_2023}. For example, while constructing the prompt
rather than simply stating the direct answer to a given problem, the answer is
formatted in a way to provide the granular steps taken to arrive at an
answer\cite{wei_chain--thought_2023} (Fig. \ref{fig:cot}). This method is
particularly useful in facilitating fidelity in multi-step arithmetic problems
along with providing interpretable insight into reasoning within the
LLM\cite{wei_chain--thought_2023}.

In addition to formatted user prompts, CoT reasoning provides a useful avenue to
monitor large language model outputs for potential exploits that may produce
misaligned behavior output\cite{baker_monitoring_2025}. This has been shown with
the monitoring of verbose CoT outputs from larger models (i.e. o3-mini) using
weaker models (i.e. GPT-4o) to prevent reward hacking
schemes\cite{baker_monitoring_2025}. For example, \textit{Baker et
al.}\cite{baker_monitoring_2025} highlights an example where by monitoring the
CoT of a model's trajectory using a separate agent, a reward hacking scheme of
modifying unit tests to always pass is thwarted. This proves useful in directing
the model to complete tasks using the correct approach rather than choosing the
simpler, often incorrect, approach. However, the authors have found that given
too much optimization the model can learn hide its intent within the CoT
producing avenues where in which hallucination can
occur\cite{baker_monitoring_2025, openai_gpt-oss-120b_2025}.

\subsubsection{Zero-Shot}
\label{sec:zero_shot}
With the increasing size of Large Language Models, Zero-Shot reasoning has been
shown to be sufficient in eliciting deeper thought responses without the need
for step-by-step examples.\cite{kojima_large_2023} Rather, a simple addition to
the prompt such as ``Let's think step by step" would be sufficient in
encouraging the model to produce a more well formed
answer.\cite{kojima_large_2023} This enables a minimalist approach to probe for
complex reasoning with the large language model leveraging the large corpus of
data that the model has been trained on \cite{kojima_large_2023,
brown_language_2020}.

This type of reasoning is often baked into the large language model with a
fine-tuning method called Instruction Tuning (Section \ref{sec:it})
\cite{wei_finetuned_2021}. Wei et al.\cite{wei_finetuned_2021} utilizes this
technique to further train large language models with Natural Language
Instruction templates to better elicit stronger inference capability from the
model. In the developed 137B parameter Finetuned Language Net (FLAN) model, the
authors find that FLAN's zero-shot performance outperformed the zero-shot
performance of the 175B parameter GPT-3 in over 80\% of evaluations
\cite{wei_finetuned_2021}.

\subsubsection{ReAct}
\label{subsec:reasoning_architecture_react}
ReAct (Reason + Act) is a general paradigm that combines reasoning and actions
within the large language model to utilize feedback to make informed choices for
the next set of actions.\cite{yao_react_2022} By utilizing prompt based approach
to navigating through an action space, ReAct is able to update its current
policy by reasoning over it's current context and
observations.\cite{yao_react_2022} This is achieved by decomposing a given task
into a smaller set of steps similar to the Chain-of-Thought
process\cite{yao_react_2022, wei_chain--thought_2023}. At a given timestep
($t$), each step consists of a language space action ($\hat{a}_t$) which Yao et
al.\cite{yao_react_2022} refer to as \textit{thought} or \textit{reasoning
trace}, an environmental action ($a_t$) such as a tool call, and an observation
($o_t$) which is the result of action ($a_t$). The LLM generates a policy
($\pi(a_t|c_t)$) for the next action ($a_t$) given the current context ($c_t$)
which consists of all actions and observations from previous timesteps. A
language space action or aforementioned \textit{thought} is performed to update
the context ($c_{t+1} = (c_t, \hat{a}_t)$) allowing for dynamic policies which
can be adjusted with feedback\cite{yao_react_2022}.

Each step is composed of a ``Thought", ``Action" and ``Observation" which the
LLM is prompted to complete.\cite{yao_react_2022} The ``Thought" is the language
space action that the LLM produces to create the updated context from the
existing context space after both an Action and Observation are
performed.\cite{yao_react_2022} ``Actions" are then performed by parsing the
subsequent output from the LLM to search for tools that match a specific syntax
(i.e. \texttt{\textbf{search}[entity]}, \texttt{\textbf{lookup}[string]}, or
\texttt{\textbf{finish}[answer]}). The respective function is then executed with
the provided argument producing an ``Observation" which is then appended to the
context before moving onto the next step. This ``Thought", ``Action" and
``Observation" process is repeated until either the LLM produces an ``Action"
consisting of \texttt{\textbf{finish}[answer]} or an iteration limit is
reached.\cite{yao_react_2022} During this process, the CoT reasoning is visible
throughout each step providing transparency into the mechanisms used to
construct the final answer.\cite{yao_react_2022}

\subsection{Domain Adaptation}
Large Language Models are pretrained on a corpus of available data with
modalities in natural language text \cite{brown_language_2020,
openai_gpt-oss-120b_2025}, general images \cite{alayrac_flamingo_2022,
li_blip_2022, radford_learning_2021}, and video sequences
\cite{alayrac_flamingo_2022, arnab_vivit_2021}. Pretraining these models on a
diverse set of data builds general knowledge and reasoning capabilities useful
for generating comprehensible responses for user queries
\cite{brown_language_2020}. The general knowledge embedded into the model from
pretraining can be leveraged and further adapted to specialize in specific
applications or downstream tasks through methods such as domain adaptive
pretraining, such as those in biology, chemistry, and other fields
\cite{gururangan_dont_2020, ock_catalyst_2023, lee_biobert_2020,
beltagy_scibert_2019}. Low-Rank Adaptation is a common approach to injecting
this domain knowledge into the large language model without retraining all of
the model parameters, effectively utilizing available resources and optimizing
on memory and computation \cite{hu_lora_2021}. Through supervised fine-tuning,
the behavior of the large language model can be adjusted to further align with
its downstream application via methods such as instruction tuning
\cite{wei_finetuned_2021, liu_visual_2023}.

\subsubsection{Domain-Adaptive Pretraining}
\label{sec:dapt}

Domain-Adaptive Pretraining (DAPT) within large language models continues the
next token prediction self-supervised training process by utilizing a smaller,
yet focused set of data \cite{gururangan_dont_2020, ke_continual_2023,
xu_bert_2019, ke_adapting_2022}. For instance, the subsequent dataset for DAPT
could include text from research papers \cite{gururangan_dont_2020}, textual
representations of atoms \cite{ock_catalyst_2023}, or multi-domain scientific
papers \cite{beltagy_scibert_2019}. Gururangan et al.
\cite{gururangan_dont_2020} explores the application of DAPT in the domains of
BioMed (2.68M papers) \cite{lo_s2orc_2020}, CS (2.22M papers)
\cite{lo_s2orc_2020}, News (11.90M articles) \cite{zellers_defending_2019}, and
Reviews (2.475M reviews)\cite{he_ups_2016} on the RoBERTa
\cite{liu_roberta_2019} model for a single pass on each dataset. The authors
observe that DAPT generates improved responses over the base RoBERTa
\cite{liu_roberta_2019} model in all domains, particularly in the BioMed, CS, and
Reviews domain showcasing the benefits such as an approach has when the source
domain of the model is distance from the target domain of the model.


\subsubsection{Low-Rank Adaptation}
\label{sec:lora}

\begin{figure}
    \centering
    \includegraphics[width=0.60\textwidth]{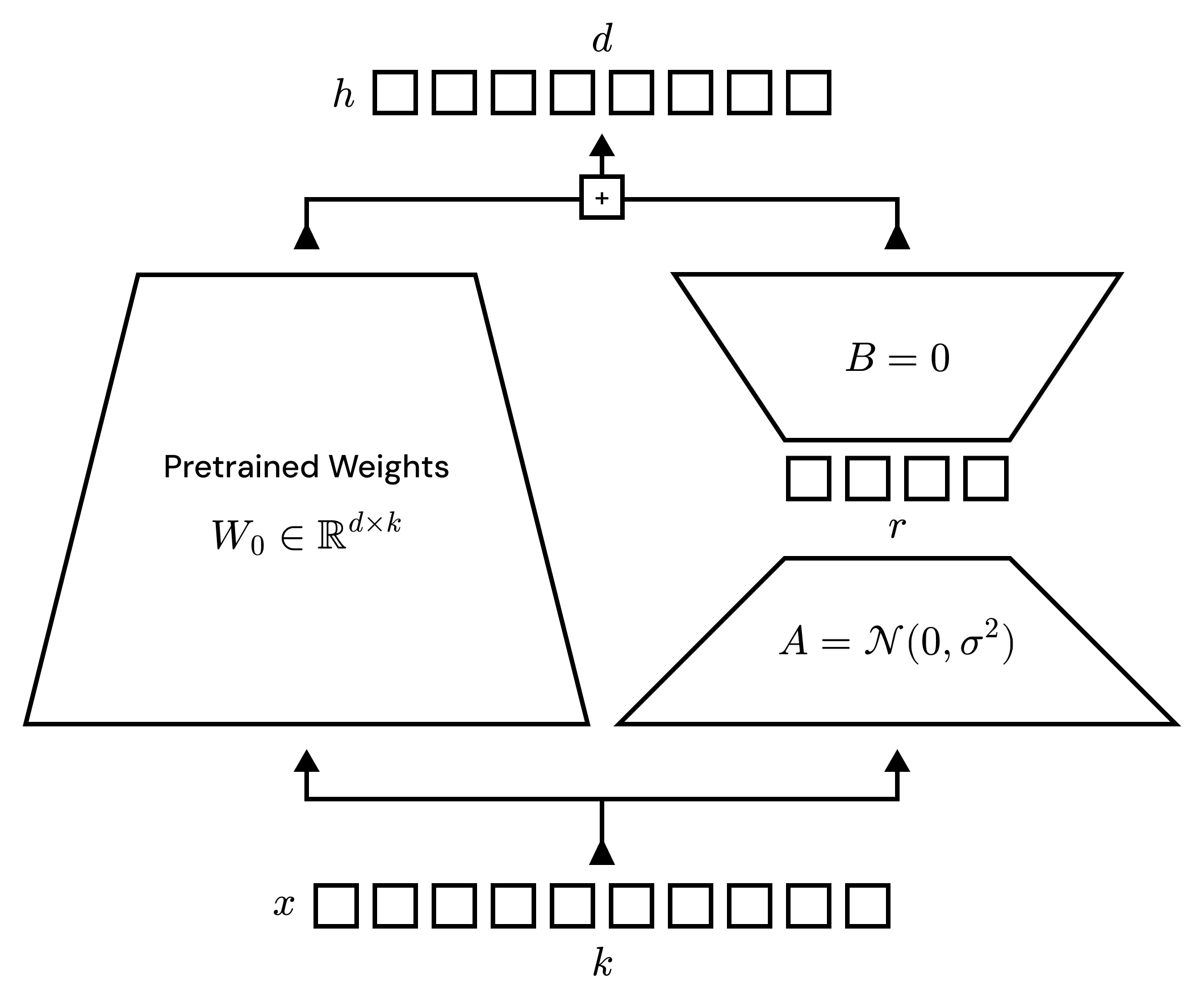}
    \caption{
        Original figure by Hu et al. \cite{hu_lora_2021} diagrams LoRA for a
        square pretrained weight matrix ($d \times d$). Figure is adapted to
        showcase an example of a pretrained weight matrix with different input
        and output dimensions ($d \times k$).
    }
    \label{fig:lora}
\end{figure}

With the increasing scale of parameters in large language models, adjusting each
parameter via fine-tuning becomes prohibitively expensive \cite{hu_lora_2021}.
As of writing, many popular large language models such as GPT-3 (175
B)\cite{hu_lora_2021}, GPT-OSS (20B and 120B)\cite{openai_gpt-oss-120b_2025},
Llama 4 (109B)\cite{arxiv_llama_2026} surpass 100 billion parameters, with
expectations to scale to over 1 trillion trainable
parameters\cite{fedus_switch_2022}. Pretraining alone for these models can take
upwards of several months and retraining each model to a specific application
evolves from an inconvenient task to an infeasible endeavor \cite{hu_lora_2021}.
This growing inaccessibility of retraining all parameters of large language
models to a specific domain establishes need for a more efficient method approach
to fine-tuning.
To this end, consideration towards the number of effective parameters is
investigated as adjusting just these parameters would be enough to sufficiently
adapt the large language model to a specific domain
\cite{aghajanyan_intrinsic_2020, li_measuring_2018, hu_lora_2021}. This is
referred by the intrinsic dimension, providing a measurement for the minimum
number of parameters that is necessary for a model to produce satisfactory
responses to an objective function \cite{aghajanyan_intrinsic_2020,
li_measuring_2018}. With this, Aghajanyan et al. was able to achieve 90\% of the
expected performance on a sematic equivalency binary classification task on the
RoBERTa model \cite{liu_roberta_2019} by training only a select 200 parameters.
Low-Rank Adaption (LoRA) is an approach that can reduce the number of trainable
parameters by up to a factor of 10,000 with only just a third of the training
memory requirement \cite{hu_lora_2021}. Here (Figure \ref{fig:lora}), the
gradients of the pretrained weight matrix $W_0$ are kept frozen the accumulated
gradients $\Delta W$ are represented by their low-rank decomposition $BA$
($\Delta W = BA$). $B \in \mathbb{R}^{d \times r}$ and $A \in \mathbb{R}^{r
\times k}$ \cite{hu_lora_2021}. Both $A$ and $B$ are dense layers which contain
trainable parameters, $d$ is the dimension of the transformer layer of the
model, $k$ is the input dimension of the weight matrix, and $r$ is the rank that
satisfies the condition $r < \text{min}(d, k)$ \cite{hu_lora_2021}. In the
adapter, the weights for $A$ are initialized using a gaussian distribution $A =
\mathcal{N}(0, \sigma^2)$ with weights for B set to 0 \cite{hu_lora_2021}.

Hu et al.\cite{hu_lora_2021} applies this approach only to within the self
attention portion of GPT-2 ($W_q$, $W_k$, $W_v$, and $W_o$). Within the
backwards pass only the adapter weights are updated, original pretrained weights
are left frozen \cite{hu_lora_2021}. The forward pass is modified from $h=W_0x$
to $h = W_0x+\Delta W$, adding the results from the trained adapter to the
output of the frozen weights \cite{hu_lora_2021}. For inference, the pretrained
weights and the adapter weights can be added together $W = W_0 + BA$, removing
any introduction of latency and furthermore can be easily switched out for a
different adapter at minimal overhead cost \cite{hu_lora_2021}.

\subsubsection{Instruction Tuning}
\label{sec:it}

Instruction Tuning (IT) is a type of fine tuning which further aligns the LLM to
produce better question answering responses \cite{wei_finetuned_2021,
liu_visual_2023}. This process utilizes labeled datasets in a supervised
learning environment to optimize the prediction performance of the model to the
corresponding label for a specific input \cite{wei_finetuned_2021,
liu_visual_2023}. Wei et al. \cite{wei_finetuned_2021} investigates the
effectiveness of IT on zero-shot prompts to understand whether the model will
exhibit unseen task performance when fine-tuning on a collection tasks. The
authors find that performance improvements scale with the number of IT tasks and
also denote that a specific model parameter size is necessary to realize these
gains \cite{wei_finetuned_2021}. For models under a specific parameter size
(8B), IT has been observed to hinder task performance potentially due to the
limited model capacity which instruction tuning then completely consumes
\cite{wei_finetuned_2021}. The IT FLAN model was shown to outperform the GPT-3
model in a number of unseen tasks and showcases the potential for even larger
LLMs to perform well in zero-shot question answering prompts
\cite{wei_finetuned_2021}.

This practice can be extended to visual domain with Visual Instruction Tuning
(VIT) \cite{liu_visual_2023}. In work by Liu et al. \cite{liu_visual_2023}, the
authors take a multi-modal approach to IT in their Large Language and Vision
Assistant (LLaVA) model achieving an accuracy of 92.53\% when evaluating on the
Science QA dataset \cite{lu_learn_2022}. The images obtained
\cite{lin_microsoft_2015} from COCO include both caption and object localization
which are then utilized in generating the three IT formats:
\textit{Conversation}, \textit{Detailed Description}, and \textit{Complex
Reasoning} \cite{liu_visual_2023}. The \textit{Conversation} format (58K
samples) utilizes only the caption text to generate 3 question and answer
responses with the assistance of GPT-4 for each image. The \textit{Detailed
description} (23K samples) utilizes the questions from the previous format to
also prompt GPT-4 for a comprehensive description of the image
\cite{liu_visual_2023}. Lastly, the \textit{Complex Reasoning} (77K samples)
format elicit more step-by-step answers to a given question
\cite{liu_visual_2023}.

Liu et al.\cite{liu_visual_2023} chose the Vicuna\cite{chiang_vicuna_2023} model
(13B) to perform their IT experiments upon. Within their ablation studies it was
found that without any instruction tuning, the model performed poorly within all
categories, demonstrating the effectiveness of IT\cite{liu_visual_2023}. By just
adding in \textit{Conversation} to the IT dataset, significant improvement gains
are observed with evaluations across all three IT formats by a minimum of 30
points and the incorporation of the entire IT dataset increases the score by 50
points across the board \cite{liu_visual_2023}. When compared to GPT-4 the
multi-modal capabilities of LLaVA produces a relative score of around 85\% on
instruction-following dataset examples \cite{liu_visual_2023}.

\section{Methodology}

A collection of large language models (Gemma 3, Qwen 3, Gemma 4) are adapted to
the field of additive manufacturing using domain adaptive pretraining (Section
\ref{sec:dapt}) and instruction tuning (Section \ref{sec:it}). The 12 billion
parameter instruction tuned variant of the Gemma 3 model \cite{team_gemma_2025}
is used as the base as it provides a multi-modal architecture capable of
performing inference upon natural language and visual inputs. The base modes are
trained on both the text and images extracted from various open-access additive
manufacturing journal articles including \textit{Journal of Additive
Manufacturing}, \textit{Additive Manufacturing Letters}, \textit{Journal of
Manufacturing Processes}, and \textit{Rapid Prototyping Journal}. Visual
instruction tuning examples are created from the extracted data using the 120
billion parameter GPT-OSS model which along with the extracted text and images
are uploaded and hosted HuggingFace \cite{peter_pak_additivellm2-oa_nodate}. The
development of the model is evaluated with the
\texttt{Additive-Manufacturing-Benchmark} dataset consisting of general
knowledge questions regarding additive manufacturing, visual identification
tasks, and other data based prediction tasks.

\subsection{Models}
The adapted models are expected to provide enhanced domain expertise within the
field of additive manufacturing. To achieve this each LLM will also need to
utilize a vision transformer in order to function beyond the scope of natural
language as additive manufacturing tasks are often a multi-modal challenge.
Vision based approaches have been utilized within the space of additive
manufacturing for tasks such as porosity prediction \cite{pak_thermopore_2024,
bostan_accurate_2025}, melt pool estimation \cite{ogoke_deep_2024}, and build
monitoring \cite{jadhav_llm-3d_2025}. Open weight LLMs from several frontier
labs were considered for this work, these include those from OpenAI
\cite{openai_gpt-oss-120b_2025}, Meta \cite{arxiv_llama_2026}, and Google
\cite{team_gemma_2025}. The 20 billion parameter variant of GPT-OSS
\cite{openai_gpt-oss-120b_2025} offers a promising foundation, however its
architecture is constrained to the medium of text. Llama 4 Scout
\cite{arxiv_llama_2026} is multi-modal LLM and utilizes a mixture of elements
architecture with 17 billion active parameters and 16 experts. Although only 17
billion parameters are active at a given time, 109 billion parameters still need
to be loaded into memory just for inference, exceeding the hardware capacity of
this experimental setup. The Gemma family of models \cite{team_gemma_2025,
noauthor_gemma_nodate} provides a suitable balance between functionality and
memory consumption while also accepting inputs within the vision space. Other
models such as the Qwen3 series of models have also demonstrated significant
capability in both the vision and reasoning space\cite{yang_qwen3_2025}. Three
different models were selected to use as the base model: Gemma 3 12B
(pre-trained and instruction tuned), Gemma 4 31B (instruction tuned only), and
Qwen 3 8B (instruction tuned only).

\subsubsection{Gemma 3}
The Gemma 3 family of models \cite{team_gemma_2025} offers a wide selection of
models that can operate on consumer grade hardware with comparable performance
(in the 27B variant evaluated with the Chatbot Arena \cite{chiang_chatbot_2024}
rating system) to other models such as DeepSeek-R1
\cite{deepseek-ai_deepseek-r1_2025}, Qwen2.5-Max \cite{qwen_qwen25_2025},
ChatGPT-4o, and Claude 3.7 Sonnet. Four different variants of the Gemma 3 model
are available those being 1B, 4B, 12B, and 27B parameter configurations, all
capable of multi-modal text and image inference with the exception of the 1B
parameter variant \cite{team_gemma_2025}. The Gemma 3 models utilize a
decoder-only transformer architecture and the 400M variant of the SigLIP vision
encoder (frozen) \cite{zhai_sigmoid_2023} for handling visual inputs, composing
separate vision and language stacks \cite{team_gemma_2025}. When compared to other
variants within the Gemma 3 family, the 12B IT variant performs near to that of
the 27B IT and significantly better than the immediately lower 4B IT variant on
various general reasoning and understanding benchmarks \cite{team_gemma_2025,
wang_mmlu-pro_2024, jain_livecodebench_2024, rein_gpqa_2023, wei_measuring_2024,
singh_global_2025, yue_mmmu_2024}.

Pretraining of the Gemma 3 family of models was achieved through a process known
as knowledge distillation \cite{hinton_distilling_2015}, a supervised learning
technique where a smaller ``student" model is trained to mimic the behavior of
the more capable ``teacher" model. Specific to the 12 billion parameter variant
of Gemma 3, a total of 12 trillion tokens were used during the pretraining
process consisting of images, text, and multilingual data to improve language
coverage \cite{team_gemma_2025}. Post training procedures also utilize knowledge
distillation for instruction tuning tasks along with improved variations of
reinforcement learning techniques such as BOND \cite{sessa_bond_2024}, WARM
\cite{rame_warm_2024}, and WARP \cite{rame_warp_2024}.

\subsubsection{Qwen3}
The Qwen3 family of models utilized a diverse multi-lingual dataset spanning
across 119 languages, totalling to approximately 36 trillion tokens for its
three stage pretraining process. The first of the three stages is the general
stage concerned with embedding general world knowledge utilizing a sequence
length of 4,096 tokens. The second stage involves pre-training for reasoning
tasks utilizing a corpous of STEM related data and coding data. The final stage
of pre-training focuses upon the long context performance, extending the
sequence length of tokens to 32,768 and selecting text samples predominantly
consisting of longer token sequences \cite{yang_qwen3_2025}. The 8 billion
parameter variant of the Qwen3 is a dense 36 layer model with a context length
of 128K tokens. In benchmarks, the Qwen3-8B surpassed Gemma-3-12B in MMLU
\cite{wang_mmlu-pro_2024}, GPQA \cite{rein_gpqa_2023}, and other reasoning
benchmarks \cite{yang_qwen3_2025}. This model was also selected as a base model
to obtain insight on the domain adaptation process when applied to a model
similar to Gemma-3-12B. 

\subsubsection{Gemma 4}
The Gemma 4 class of models builds upon the foundation of the previous Gemma 3
models providing options for either dense and Mixture-of-Experts (MoE) based
architectures exhibiting improvements in reasoning, multimodality, device
optimization, and context length \cite{noauthor_gemma_nodate}. The selected
Gemma 4 31B dense model provides a context window of 256K tokens, supports both
text and image modailities, and has a total of around 550M vision encoder
parameters \cite{noauthor_gemma_nodate}. Deviating from the previous training
process for Gemma 3, only the instruction tuned variant of this model was
utilized as it was observed that building from the pre-trained base does not
reach the capability of the even the base instruction tuned variant
(\ref{sec:gemma_3_pt}). This much larger parameter model was chosen to explore
the performance of the domain adaptation process when unconstrained by parameter
limitations.

\subsection{Dataset}
\begin{figure}
    \centering
    \includegraphics[width=\textwidth]{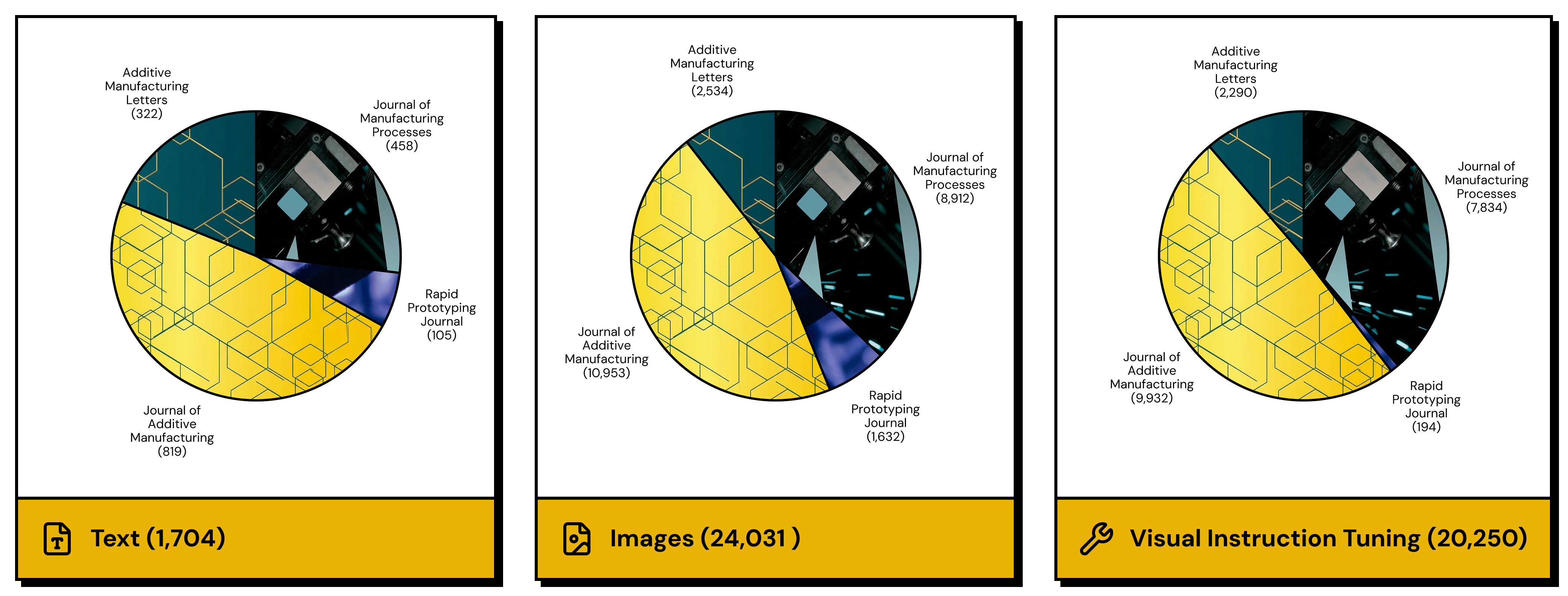}
    \caption{
        Composition of each dataset configuration by journal for \textbf{(left)}
        text, \textbf{(center)} images, and \textbf{(right)} visual instruction
        tuning (VIT). The reduced \textit{Rapid Prototyping Journal} slice in
        the VIT configuration \textbf{(right)} is due to the smaller number of
        figure and caption pairs found within that journal during the data
        extraction process.
    }
    \label{fig:journal_pie_charts}
\end{figure}

The multi-modal dataset hosted on
HuggingFace\cite{peter_pak_additivellm2-oa_nodate}) created for domain
adaptation was sourced from four different peer reviewed journals:
\textit{Journal of Additive Manufacturing}, \textit{Additive Manufacturing
Letters}, \textit{Journal of Manufacturing Processes}, and \textit{Rapid
Prototyping Journal}. Text, images, and visual instruction tuning (VIT) examples
compiled for this dataset utilized all articles (1,704 total) published within
these journals under the open-access license up to February of 2026. Among the
keywords associated with these articles, Laser Powder Bed Fusion as the most
common followed by other processes such as material extrusion, directed energy
deposition, and vat photopolymerization (Fig. \ref{fig:keywords_top10}).
Composition (text, images, VIT) of each configuration by journal with
\textit{Journal of Additive Manufacturing} accounting for a majority of each
configuration followed by \textit{Journal of Manufacturing Processes} (Fig.
\ref{fig:journal_pie_charts}). Text was obtained from these 1,704 articles and
amounts to around 29 million tokens with \textit{Rapid Prototyping Journal} and
\textit{Journal of Additive Manufacturing} exhibiting the most distinct
vocabulary from one another (Fig. \ref{fig:vocabulary_overlap} and Fig.
\ref{fig:ngrams}).

All articles were downloaded from their respective journals in the \texttt{.pdf}
format and using the \texttt{PyMuPDF}\cite{mckie_pymupdfpymupdf_2026} library,
the relevant text and images were extracted from each file. The data extraction
process for each assumes a consistent content structure in order to parse
attributes such as authors, keywords, and figure captions. In the few cases
these attributes are not parsed properly author names are incomplete, keywords
are left out, or figures captions are left empty. This issue mostly affects a
small number of the early articles from each journal as the latter issues adopt
a consistent formatting.

\begin{figure}
    \centering
    \includegraphics[width=\textwidth]{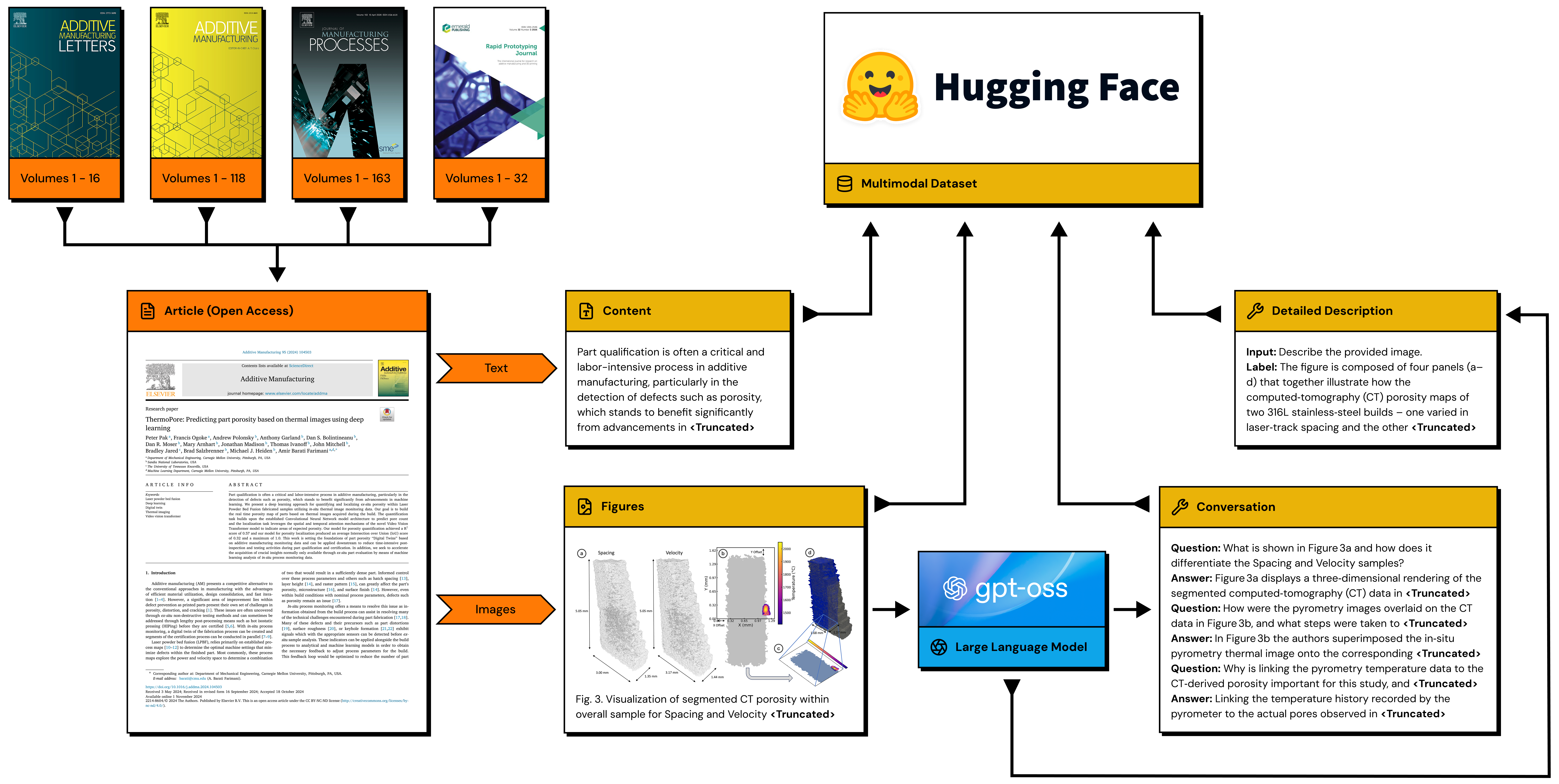}
    \caption{
        Content (text) and figures (images) are extracted from their
        \texttt{.pdf} formats from open-access articles from various additive
        manufacturing journals. This data is utilized for domain adaptive
        pretraining and the figure captions are processed into question answer
        conversations and detailed descriptions using
        GPT-OSS\cite{openai_gpt-oss-120b_2025} for visual instruction tuning.
        The various data configurations are uploaded and publicly accessible via
        HuggingFace.
    }
    \label{fig:dataset}
\end{figure}

Over 24,000 images were extracted for this dataset (Fig.
\ref{fig:journal_pie_charts}) amounting to over 6 million image tokens and over
10 million text tokens for their respective captions
\cite{peter_pak_additivellm2-oa_nodate}. These images were primarily obtained
from article figures with consideration towards maintaining the association
between the caption text. These pairs would be utilized in the domain adaptive
pretraining process of the model utilizing images (Fig. \ref{fig:training}).
Following the practices outlined by Liu et al. \cite{liu_visual_2023} in their
work for visual instruction tuning, detailed descriptions and conversation
examples were generated using the extracted image captions. A local deployment
of the GPT-OSS (120B) model \cite{openai_gpt-oss-120b_2025} was used to generate
the detailed description and question-answer conversation examples using only
the figure's caption as input. If a figure's image-caption pair was not properly
extracted, its VIT examples were not generated. This provides an explanation why
\textit{Rapid Prototyping Journal} occupies a smaller fraction of the VIT
dataset configuration when compared to the other configurations of text or
images (Fig. \ref{fig:journal_pie_charts}). Around 20,000 VIT examples were
compiled amounting to a total of around 12 million text tokens and 5 million
image tokens \cite{peter_pak_additivellm2-oa_nodate}. In total the dataset
consists of around 57 million tokens: 45 million of which are text and the other
11 million that are image.

\subsection{Training}
\begin{figure}
    \centering
    \includegraphics[width=\textwidth]{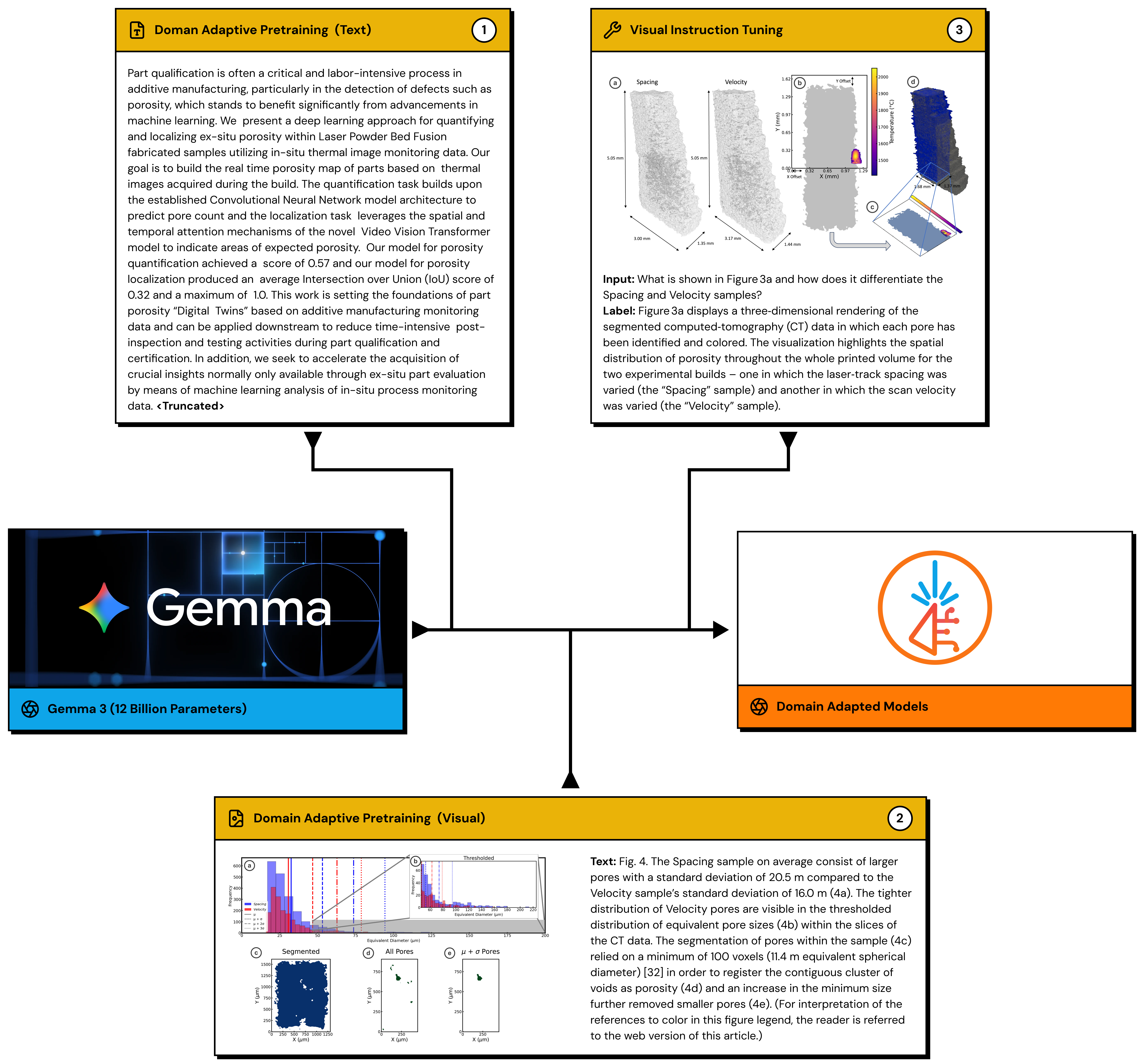}
    \caption{
        Gemma 3 12B variants (pretrained only or instruction tuned) are adapted
        to the domain of additive manufacturing using LoRA in a 3 step training
        process. \textbf{(1)} Extracted text from the open-access articles is
        used to train language model attention weights for next token
        prediction. \textbf{(2)} Extracted figures along with their respective
        captions are utilized to train the vision tower attention weights while
        keeping the language model weights frozen. \textbf{(3)} Visual
        instruction tuning post training on all attention weights is performed
        using question-answer pairs along with detailed image descriptions.
    }
    \label{fig:training}
\end{figure}

Training for the various models was split into three sequential stages: Domain
Adaptive Pretraining (Text), Domain Adaptive Pretraining (Images), and Visual
Instruction Tuning (Conversation and Detailed Description). This is outlined in
Figure \ref{fig:training} where the Gemma 3 \cite{team_gemma_2025} model is used
as the base for text and image based DAPT using the content extracted from
various open-access AM articles, further instruction tuned with generated
input-label pairs from the extracted figures. The \texttt{text},
\texttt{images}, and \texttt{vit} configurations of the training dataset
\cite{pak_ppak10additive-manufacturing_2025} hosted on HuggingFace is used for
each respective stage in the training process. Each of the model weights are
adapted using LoRA (Section \ref{sec:lora}) with the corresponding rank ($r=16$)
and alpha ($a=32$) configurations applied to the query, key, value, and output
projection layers within self-attention blocks. Each stage of the training
process was performed for 3 epochs on a machine consisting of 3 Nvidia A6000
GPUs for a duration of around 36 hours per stage.

Domain adaptive pretraining extends the pretraining process using unlabeled data
for the task of next token prediction (Section \ref{sec:dapt}). The first stage
of the training process applies DAPT to the language modality of the model,
adapting the corresponding attention weights utilizing the vocabulary, concepts,
and phrasing within additive manufacturing articles. Training inputs during this
stage have a split of 95\% train and 5\% validation and are provided in chunks
of 2048 tokens from the \texttt{text} configuration of the training dataset
\cite{peter_pak_additivellm2-oa_nodate}. The weights of the LLM are adapted
using LoRA and merged into the base weights for the next stage of the training
process. The second stage utilizes the figures extracted from each of the
articles to adapt the vision tower of the LLM. In this stage, the attention
weights of the previous language modality are frozen as the intent is to train
the model to build an associate between figure images and caption text.

The last stage of the training process applies the post-training technique of
instruction tuning (Section \ref{sec:it}) to fine-tune the LLM's performance in
additive manufacturing related tasks. A labeled dataset of conversation
question-answer pairs and detailed descriptions generated from the extracted
figures are used to perform supervised fine-tuning (SFT). In this stage, LoRA is
performed to adapt both the language and vision weights of the model. For the
description task, the model is prompted to provide a description of the given
input image and the loss is calculated between generated response and the ground
truth. In the multi-turn conversation task, the Gemma 3 specific
\texttt{<start\_of\_turn>} token \cite{team_gemma_2025} is utilized to generate
a question answer conversation of a given input image. Within this conversation,
3 questions are provided with the loss only computed on the response to the last
question as to allow the first 2 questions to build context within the
conversation. Both methods implement prompt masking which incentivizes the model
to generate responses to given questions. The maximum length of token outputs in
this stage is increased from 768 to 1024 to accommodate the longer responses
expected during this stage.


\subsection{Benchmarking}
\begin{figure}
    \centering
    \includegraphics[width=0.8\textwidth]{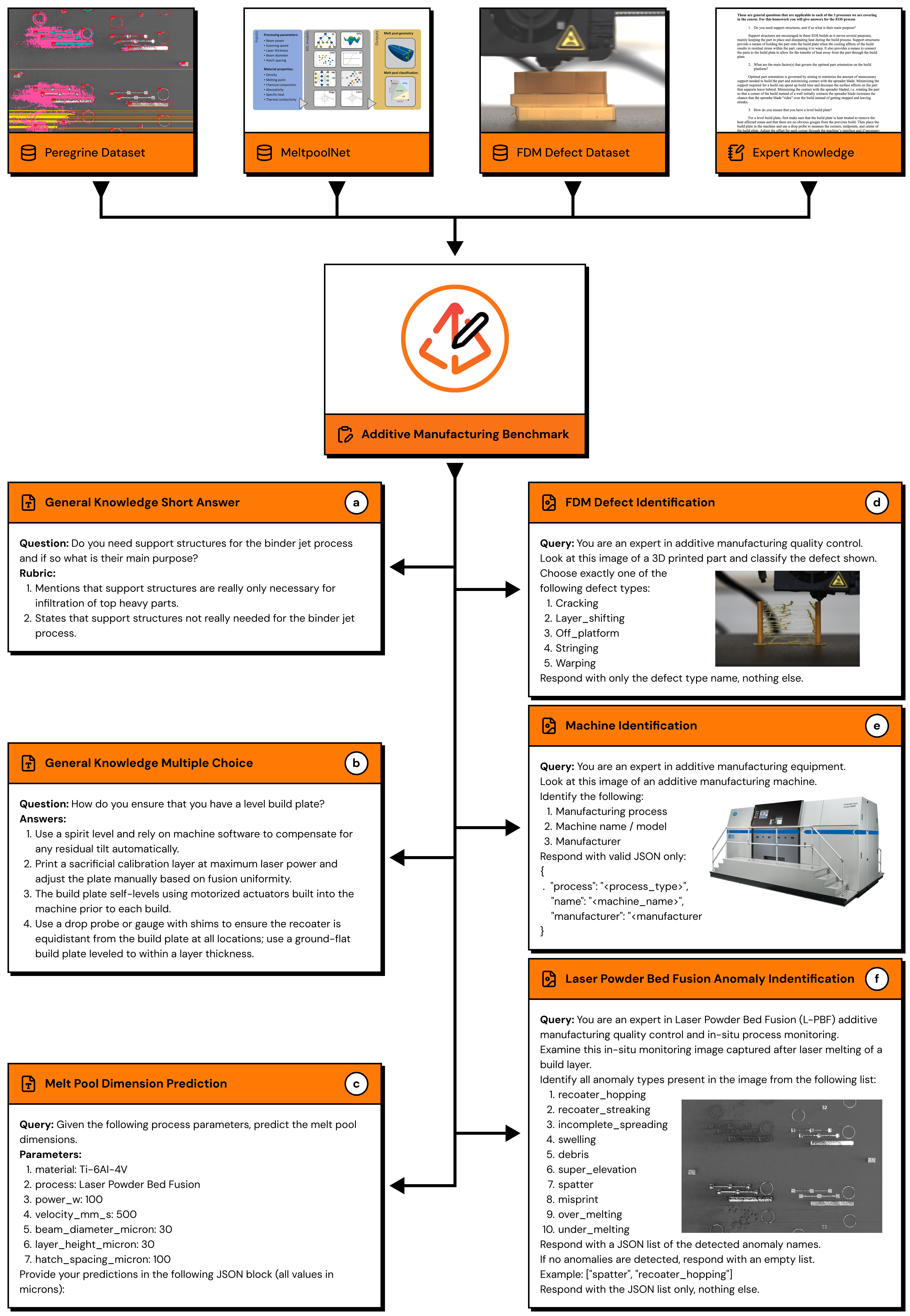}
    \caption{
        \texttt{Additive-Manufacturing-Benchmark} is a set of multi-modal tasks
        compiled from various published resources\cite{hu_real-time_2024,
        scime_layer-wise_2023, akbari_meltpoolnet_2022} used to evaluate AM
        comprehension. Language based tasks include answering general knowledge
        questions in \textbf{(a)} short answer or \textbf{(b)} multiple choice
        formats and \textbf{(c)} melt pool dimension prediction given process
        parameters. Image-based tasks including \textbf{(d)} FDM Defect
        identification, \textbf{(e)} AM machine identification, and \textbf{(f)}
        LPBF anomaly identification.
    }
    \label{fig:am_bench}
\end{figure}

To properly evaluate the capabilities of each domain adapted model, a benchmark
consisting of additive manufacturing related questions was created and
accessible through the \texttt{additive-\allowbreak manufacturing} package
\cite{pak_ppak10additive-manufacturing_2025}. \texttt{Additive-\allowbreak
Manufacturing-\allowbreak Benchmark} (Fig. \ref{fig:am_bench}) aims to provide a
comprehensive assessment in both the language and visual modalities encountered
within the field of additive manufacturing, consisting of several AM tasks
evaluating general process knowledge, defect recognition, and other adjacent
capabilities. Sourcing of the data used within these benchmarking tasks
originate from domain expert generated course materials and published datasets
\cite{hu_real-time_2024, scime_layer-wise_2023, akbari_meltpoolnet_2022}.

The tasks for general knowledge regarding additive manufacturing are delivered
in two forms: short answer and multiple choice (Fig. \ref{fig:am_bench}a and
\ref{fig:am_bench}b). In both formats there are 127 questions covering various
processes from laser powder bed fusion, binder jet, directed energy deposition,
and etc. The multiple choice format provides 4 different choices with one
correct answer and the short answer provides a rubric to evaluate short answer
questions. The short answer format utilizes the GPT-OSS 20B model
\cite{openai_gpt-oss-120b_2025} to award points based off the rubric where a
maximum score of 127 points can be achieved for this task.

The melt pool prediction task (Fig. \ref{fig:am_bench}c) provides a metric to
evaluate an LLM's familiarity with resulting melt pool dimensions associated
with a configuration of process parameters such as beam power, scanning
velocity, or material. This task utilizes experimental data from MeltpoolNet
\cite{akbari_meltpoolnet_2022} where melt pool dimensions along with the
prescribed process parameters are compiled into a cohesive dataset. This task
aims to probe the model to provide estimations of melt pool dimensions, in the
unit of microns, of the depth, length, and width given a combination of process
parameters. The predictions are evaluated using RMSE where a value closer to 0
is desired.

The Fused Deposition Modeling (FDM) defect prediction accuracy task (Fig.
\ref{fig:am_bench}d) evaluates the model's visual capability by prompting the
model to correctly assign a defect classification to a given FDM process image.
These images are obtained from Hu et al. \cite{hu_real-time_2024} and provide
defect classifications for in-situ build images include warping, stringing,
cracking, layer shift, and off platform. A total of 100 samples are provided to
the LLM and the model is evaluated on a binary task of assigning the image to
the correct defect classification.

The machines identification task (Fig. \ref{fig:am_bench}e) also evaluates the
ability of the LLM to recognize an image of a given machine. This is intended to
evaluate the model's ability to generally understand what AM machines look like
and what process each is associated with. The reasoning behind including this
task is gather insight into the incorporation of article figures into the
model's vision stack as many articles include descriptions and images of the
equipment utilized for their experiments. The model is expected to provide a
prediction for the associated process, manufacturing company, and name of the
machine provided in the input image. Half the response weight is placed on the
correct identification of the AM process associated with the machine featured in
the image with the latter two questions (name and manufacturer) accounting for
the remaining response weight.

The LPBF anomaly identification task (Fig. \ref{fig:am_bench}f) with the
Peregrine dataset\cite{scime_layer-wise_2023} aims to evaluate capability of the
LLM to recognize build anomalies within laser powder bed fusion. More
specifically, this visual identification task asks the model to classify which
anomalies exist on a given build layer after melting. These anomalies are
obtained using the Peregrine software and include classifications such as
recoater hopping, under melting, over melting, spatter, debris, and etc. Task
performance is measured using an F1 score which considers both the precision and
recall of a set of predictions ranging from a worst case of 0 and a best case of
1 (Equation \ref{eq:f1}).

\begin{equation}
\label{eq:f1}
\text{Recall} = \frac{TP}{TP + FN} \qquad \text{Precision} = \frac{TP}{TP + FP} \qquad F_1 = \frac{2 \cdot TP}{2 \cdot TP + FP + FN}
\end{equation}

\section{Results and Discussion}
The various stages of the domain adaptation process were evaluated with tasks
from the \texttt{Additive-\allowbreak Manufacturing-\allowbreak Benchmark}
performed over a set of 5 trials. Domain adapted models for the Gemma series and
Qwen series were investigated at DAPT text, DAPT images, and VIT training
iterations and compared against the base model. For all cases it was observed
that additional text pretraining and instruction tuning enabled further domain
knowledge specialization as the base model was not selected as the ``best" model
for a specific task in any of the cases except in the case of the Gemma 4 31B IT
model. Domain adaptive pretraining for images did present a noticeable decrease
in performance in specific tasks such as defect and anomaly detection throughout
all models, indicating potential loss of parameteric visual information during
the training process.

\subsection{Benchmark Results}
\begin{figure}
    \centering
    \includegraphics[width=\textwidth]{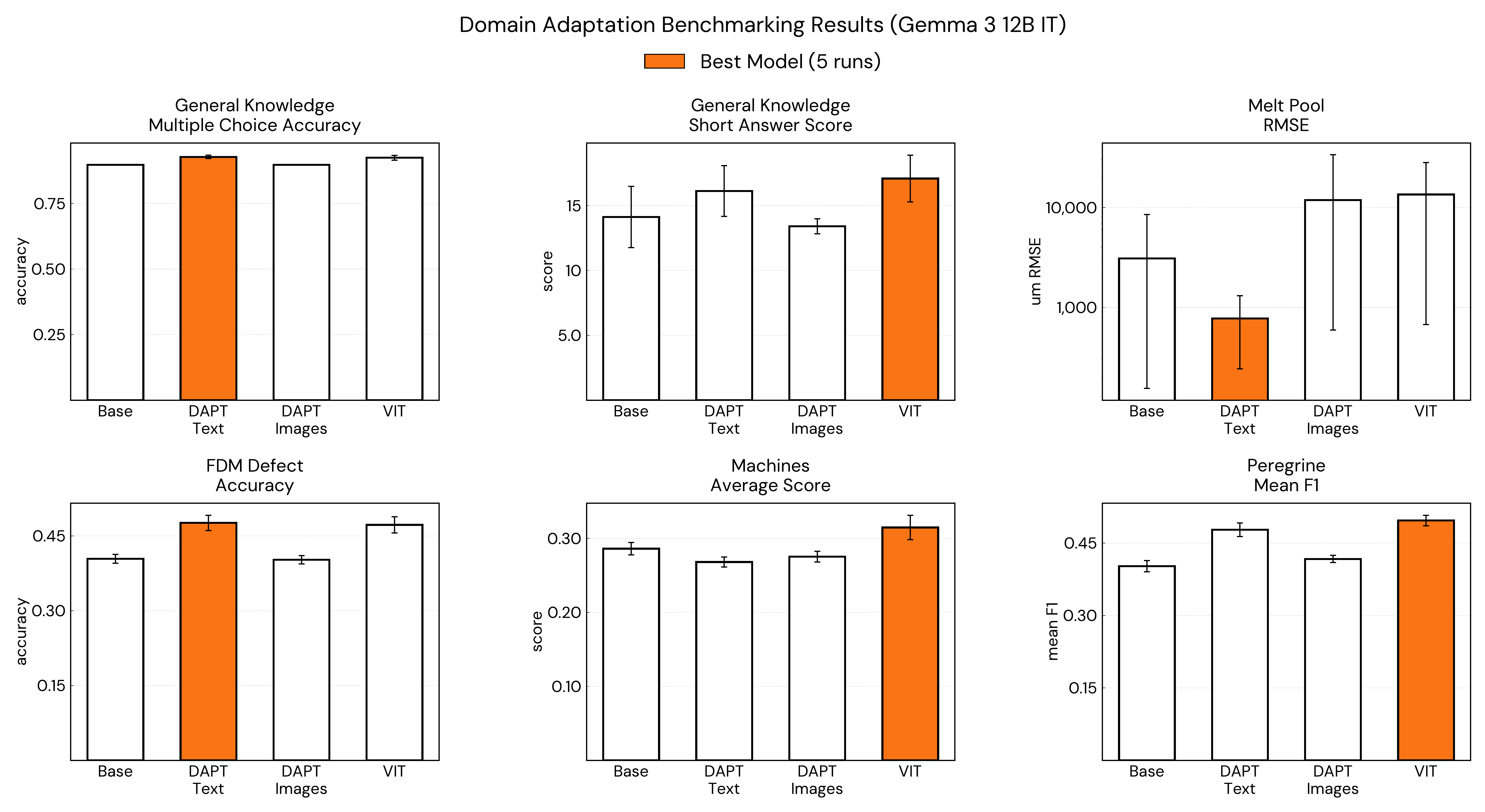}
    \caption{ 
        The visual instruction tuned version of \texttt{Gemma-\allowbreak
        3-\allowbreak 12b-\allowbreak it} exhibits the best performance in most
        image based tasks and closely follow the performance of the base model
        with only domain adaptation in text for remaining tasks.
    }
    \label{fig:gemma_3_results_it}
\end{figure}

Specific to the models using the instruction tuned base model of Gemma 3 12B (Fig.
\ref{fig:gemma_3_results_it}), these on average showcased the best performance
especially in the task of General Knowledge Multiple Choice. Within this task of
127 question, the base IT model already achieves an impressive score of around
88\% accuracy and with domain adaptive pretraining, the performance increases to
around 93\% accuracy. The short answer format of the general knowledge task
indicates a similar trend however significantly lower performance as the maximum
achievable score is 127. For the prediction of melt pool dimensions, the model
with only DAPT text training performs the best with subsequent tests for DAPT
images and VIT progressively increasing in average RMSE. The final VIT stage of
the model performs the best in visual tasks in LPBF anomaly identification and
machine identification with the exception of FDM defect identification.

\begin{figure}
    \centering
    \includegraphics[width=\textwidth]{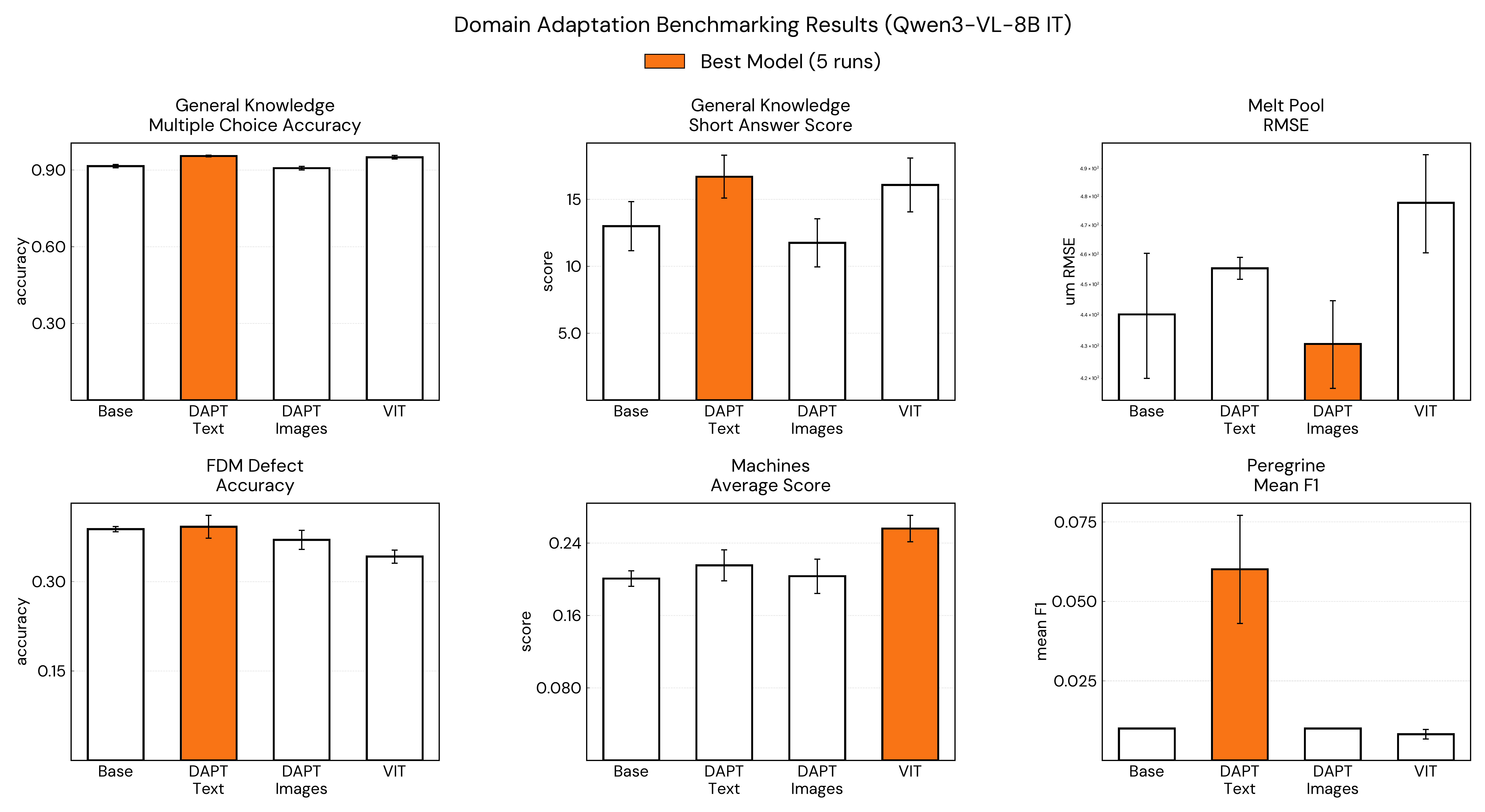}
    \caption{ 
        Domain adapted \texttt{Qwen3-\allowbreak vl-\allowbreak 8b-\allowbreak
        it} performs comparably on general knowledge tasks however under
        performs on vision based benchmarking tasks such as anomaly and defect
        detection.
    }
    \label{fig:qwen_3_results_it}
\end{figure}

The domain adpated \texttt{Qwen3-\allowbreak vl-\allowbreak 8b-\allowbreak it}
shows a similar trend in performance with general knowledge tasks however
significantly under performs on visual tasks such as those of FDM defect
detection and powder bed anomaly detection to it's Gemma 3 counterpart (Fig.
\ref{fig:qwen_3_results_it}). Although, the base \texttt{Qwen3-\allowbreak
vl-\allowbreak 8b-\allowbreak it} model outperformed the
\texttt{Gemma-\allowbreak 3-\allowbreak 12b-\allowbreak it} in various
benchmarks outlined in Qwen 3's technical documentation \cite{yang_qwen3_2025},
the smaller amount of parameters could provide an explanation to its decreased
performance on \texttt{Additive-\allowbreak Manufacturing-\allowbreak
Benchmark}.

\begin{figure}
    \centering
    \includegraphics[width=\textwidth]{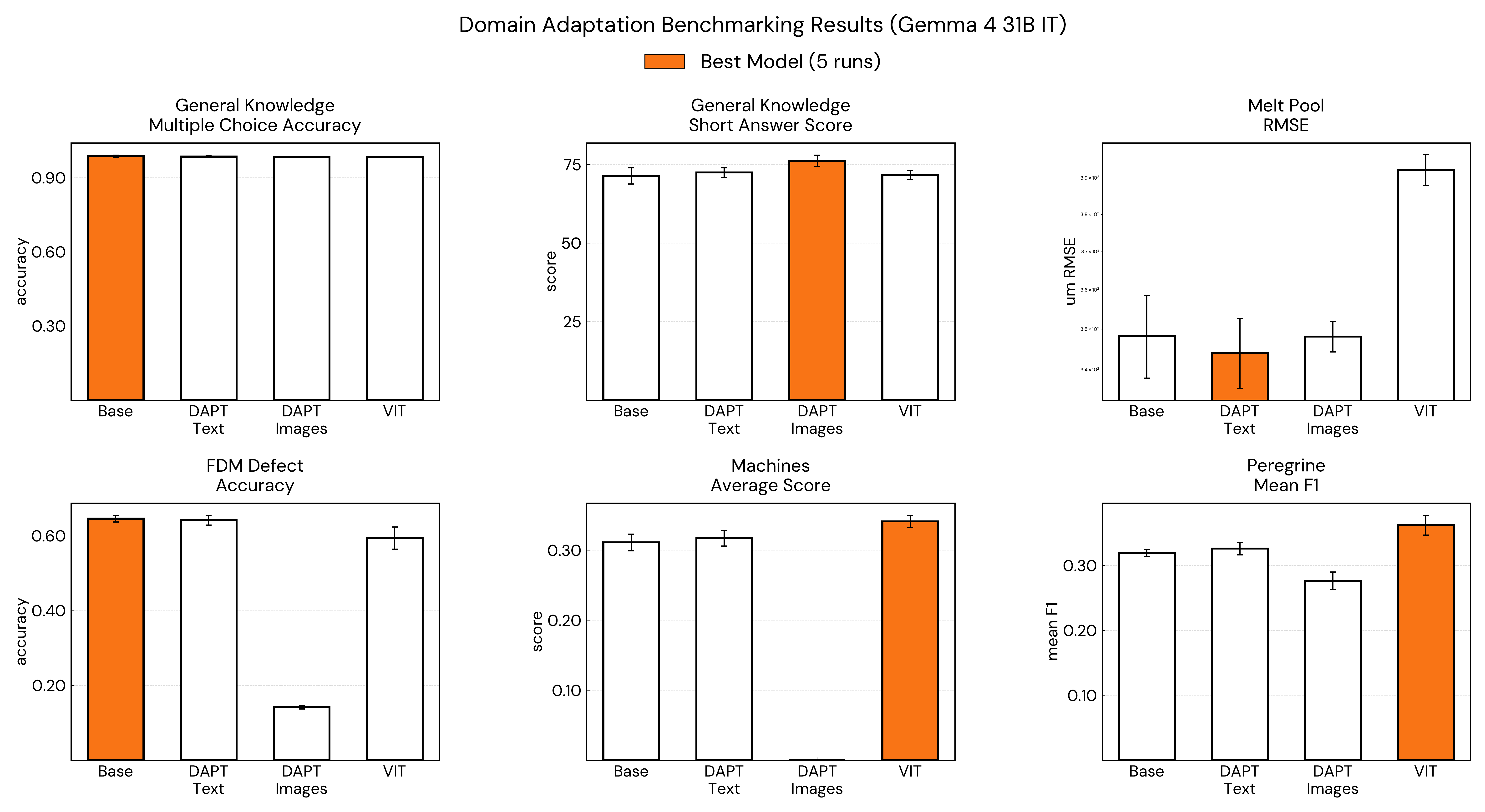}
    \caption{
       The domain adapted 31 billion parameter Gemma 4 model achieves the best
       results for general knowledge based tasks but performs comparably to
       other investigated models for image based tasks.
    }
    \label{fig:gemma_4_results_it}
\end{figure}

Of the three, Gemma 4 is the most capable but also the most resource intensive
of the compared models. This is observed with the base \texttt{gemma-\allowbreak
4-\allowbreak 31B-\allowbreak it} model that achieved the highest accuracy in
the general knowledge multiple choice and FDM defect accuracy tasks, followed
immediately by the DAPT text (Fig. \ref{fig:gemma_4_results_it}). The capability
of this model is further established in the general knowledge short answer task
where it achieves a score upwards of 75 whereas the other models only achieve a
score of around 15. However for image based tasks, the model achieves similar to
worse results than the other compared models.

\subsection{Domain Specialization Performance}
\label{sec:domain_specialization_performance}
In work by Gururangan et al. \cite{gururangan_dont_2020}, the authors perform
domain adaptive pretraining on the RoBERTA-base model (around 100 million
parameters)\cite{liu_roberta_2019} for around 1 epoch using a dataset consisting
of around 24B tokens those from BioMed (7.55B), CS (8.10B), News (6.66B), and
Reviews (2.11B). In the domain adaptation of the various models, the total token
count for text based training amount to just around 45M, considerably lower than
2B even with 3 epochs of training (135M). Yet, with this lower amount of tokens,
the final vision instruction tuned model (from \texttt{gemma-3-12b-it} base)
shows improved performance on tasks such as general knowledge short answer, LPBF
anomaly detection, FDM defect detection, and machine recognition over the base
model (Fig. \ref{fig:gemma_3_results_it}).

An explanation for this performance could be attributed to the more narrow scope
that the domain of additive manufacturing that the models apply to in comparison
to the more general domains of biology, medicine, computer science, journalism,
and consumer reviews found in Gururangan et al. \cite{gururangan_dont_2020}.
Wang et al. \cite{wang_toward_2026} used a dataset of the same size (48.5M) and
were able to achieve improved results over their base model with a similar
pretraining and fine-tuning process. More so, Junior et al.
\cite{junior_interplay_2025} observed an inverse trend with the required compute
needed for domain adaptation to the size of the model. With this, it would still
be interesting to explore the effect that a more comprehensive additive
manufacturing dataset would have to its performance.

\section{Conclusion}
In this work, the series of domain adapted models are shown to outperform their
respective base counterparts after stages of domain adaptive pretraining and
visual instruction tuning. Each model was trained on a selection of open-access
additive manufacturing articles establishing that domain adaptation is possible
on a relatively small dataset of around 45 million tokens over 3 epochs for each
stage. Evaluated on AM benchmarking tasks, the visual instruction tuned variant
of these domain adapted models (built upon \texttt{Gemma-\allowbreak
3-\allowbreak 12b-\allowbreak it}) exhibits the best performance in vision based
tasks and language tasks, achieving accuracy in general knowledge upwards of
90\%. Larger models such as the domain adapted \texttt{Gemma-\allowbreak
4-\allowbreak 31b-\allowbreak it} display greater capability in general domain
knowledge in both short answer and multiple choice form, however perform
similarly on visual tasks. With this, domain adaptive pretraining in conjunction
with instruction tuning offer an accessible method of specializing large
language models to a given domain such as additive manufacturing.

\section{Future Work}
Future work would explore extending the usage of this model with agentic
systems, enhanced datasets, and continual learning environments. The most
immediate application would investigate applying the domain adapted models into
an agentic system, evaluating the efficiency with which it is capable of making
the appropriate tool calls using its adapted domain knowledge. The performance
of this system would be evaluated in a manner similar to that of LLM-3D Print
\cite{jadhav_llm-3d_2025} where images of the process in-situ are evaluated
using the VLM component of the agentic system and the appropriate actions are
taken to address potential issues. 

Building upon the discussions in Section
\ref{sec:domain_specialization_performance} regarding domain specialization
performance, domain adaptation pretraining and instruction tuning with a more
comprehensive dataset would be worth investigating. Although the dataset used in
this work was sufficient to exhibit performance gains across many benchmarking
tasks, a larger and more encompassing dataset of additive manufacturing
processes would result greater performance gains.

Further specialization to a specific task (i.e. anomaly detection within laser
powder bed fusion) is expected to be reflected in task performance, however,
this may impact performance on other tasks. To this end, continual learning
\cite{ke_continual_2023} approaches will be further investigated and applied to
the domain adaptation process to better retain trained abilities in subsequent
training and fine-tuning cycles. This will alleviate the effects of catastrophic
forgetting \cite{french_catastrophic_1999}, building to a framework with which
the agentic system can integrate the results of its actions into the LLM through
fine-tuning.

\clearpage
\appendix

\renewcommand{\thesection}{Appendix \Alph{section}}




\section{Tokenization}
\label{sec:tokenization}
Tokenization is an essential component of the Natural Language Processing (NLP)
pipeline as it converts strings of human-readable characters into token
representations which are then embedded into vectors for the large language
model \cite{sennrich_neural_2016, alqahtani_stop_2026,
gastaldi_foundations_2025}. Raw text does not provide a suitable representation
medium for models to train upon as it commands a large vocabulary and treats
words as distinct units\cite{sennrich_neural_2016}. Thus, tokenization presents
a more efficient representation of the data to the model as an embedding
vector\cite{vaswani_attention_2023}. Tokenization methods include dividing
character strings into word and subword units (\ref{sec:snmt}) along with
indexing frequently occuring sequences detected using Byte Pair Encoding
(\ref{sec:bpe}) \cite{sennrich_neural_2016}. In order to retain positional data,
methods such as sinusoidal positional encoding\cite{vaswani_attention_2023} or
Rotary Position Embeddings (RoPE)\cite{su_roformer_2023} are added to the token
embedding vectors. By converting the tokens to vector embeddings with positional
data, the model is able to use the semantic and sequential patterns of the input
to perform next token prediction from the representations learned during
training\cite{vaswani_attention_2023, sennrich_neural_2016, su_roformer_2023}.

\subsection{Subword Neural Machine Translation}
\label{sec:snmt}
Subword Neural Machine Translation is a preprocessing method which text is
segmented into subword units, specifically useful in encoding out-of-vocabulary
(OOV) words. The approach proposed by Sennrich et al.
\cite{sennrich_neural_2016} implements an adapted version of Byte Pair Encoding
\cite{gage_new_1994} (BPE) further discussed in \ref{sec:bpe} to first generate
the pair table for frequently occurring character sequences within the train
text. This is similar to the pair table seen in the compressed output that
original BPE produces, however with slight adjustment of merging characters
rather than bytes in order to suit the application of word segmentation
\cite{sennrich_neural_2016}. Along with this, the compression routine is set to
conclude after a given number of operations rather than the original BPE process
of repeating until there are no more remaining bytes in the
text\cite{gage_new_1994}. This provides a tunable \texttt{num\_operations}
parameter which balances the frequency for complete words and subwords within
the dictionary, improving the coverage of tokens during training. This allows
for out-of-vocabulary words to be segmented into combinations of word and
subword tokens.

\subsection{Byte Pair Encoding}
\label{sec:bpe}
Byte Pair Encoding was first introduced by Philip Gage \cite{gage_new_1994} as a
method of data compression useful in memory constrained environments due to its
fast expansion routine. The compression routine of the algorithm looks for most
adjacent byte pairs that occur most frequently within a given pass and replaces
the pair with a byte that doesn't already exist within the data. This repeats
until there is either no more frequent byte pairs or there are no more remaining
unused bytes\cite{gage_new_1994}. The expansion routine is performed over a
single pass over the input file, where byte literals are passed directly to the
output buffer and byte pairs are pushed onto a stack. Within each iteration, if
the stack contains data the byte there is used as the next input byte, otherwise
the next input byte is obtained from the input file.
This is the content for the first appendix.
\clearpage

\section{Chain of Thought Prompting}
\begin{figure}
    \centering
    \includegraphics[width=0.75\textwidth]{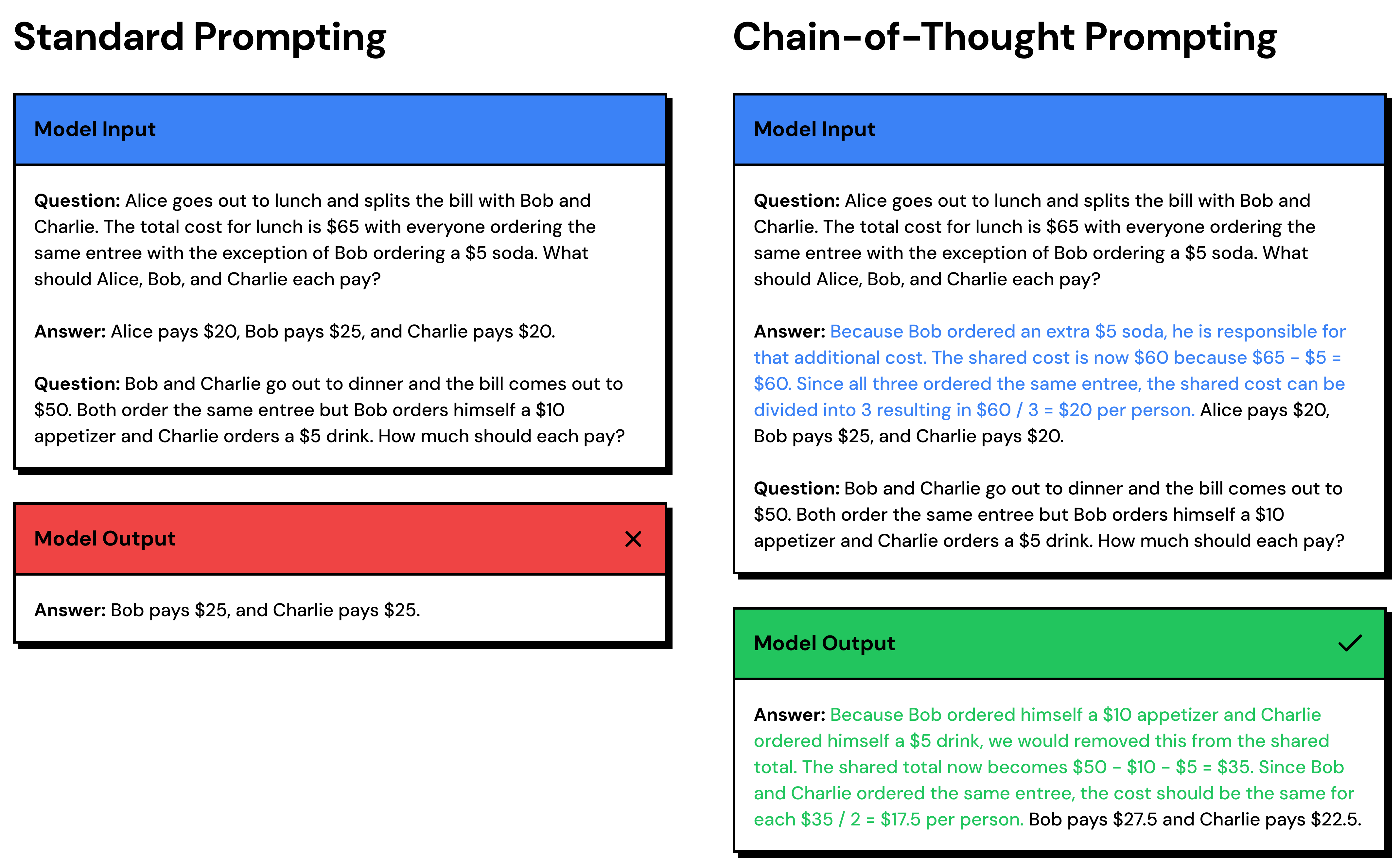}
    \caption{
    Hypothetical comparison of outputs between standard prompting process to
    that of the chain-of-thought reasoning process.
    }
    \label{fig:cot}
\end{figure}
\clearpage

\section{Additional Dataset Information}
\label{sec:additional_dataset_Information}

\begin{figure}
    \centering
    \includegraphics[width=\textwidth]{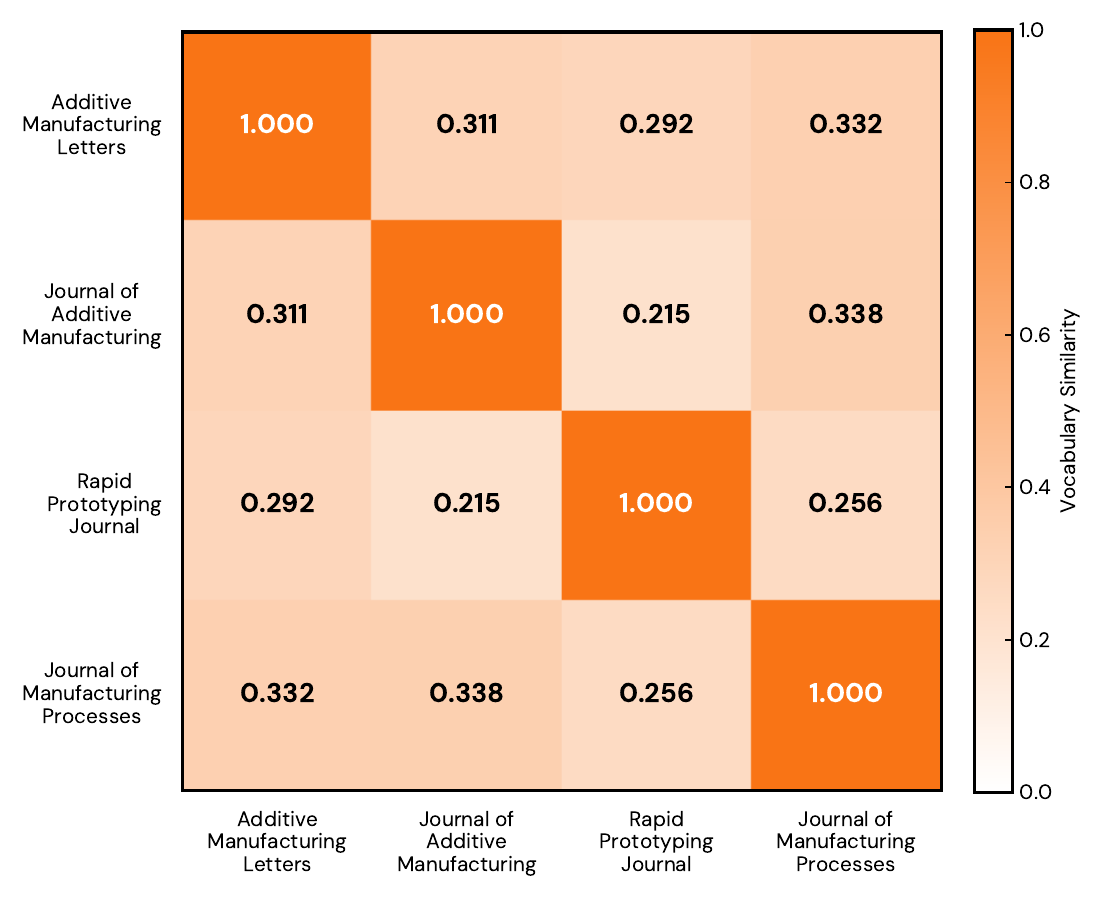}
    \caption{
        Plot indicates vocabulary similarity between the various open-access
        articles categorized by their respective journal used in the dataset.
        \textit{Additive Manufacturing Letters} exhibits the highest vocabulary
        similarity between all journals with the lowest vocabulary similarity
        seen between \textit{Rapid Prototyping Journal} and \textit{Journal of
        Additive Manufacturing}.
    }
    \label{fig:vocabulary_overlap}
\end{figure}

\begin{figure}
    \centering
    \includegraphics[width=\textwidth]{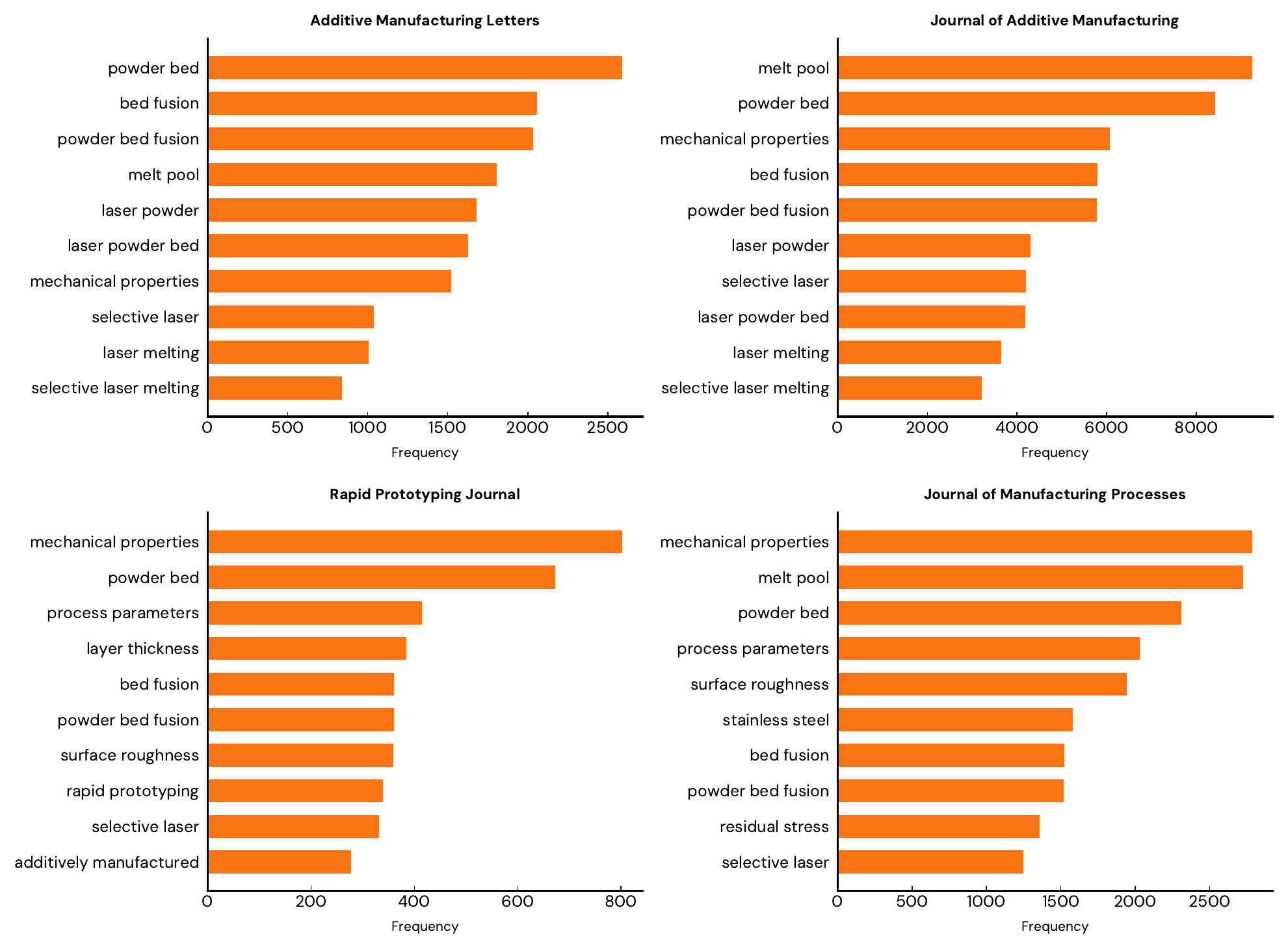}
    \caption{
        Common n-grams ordered by frequency from each journal with common stop
        words and journal information filtered out. 
    }
    \label{fig:ngrams}
\end{figure}

\begin{figure}
    \centering
    \includegraphics[width=\textwidth]{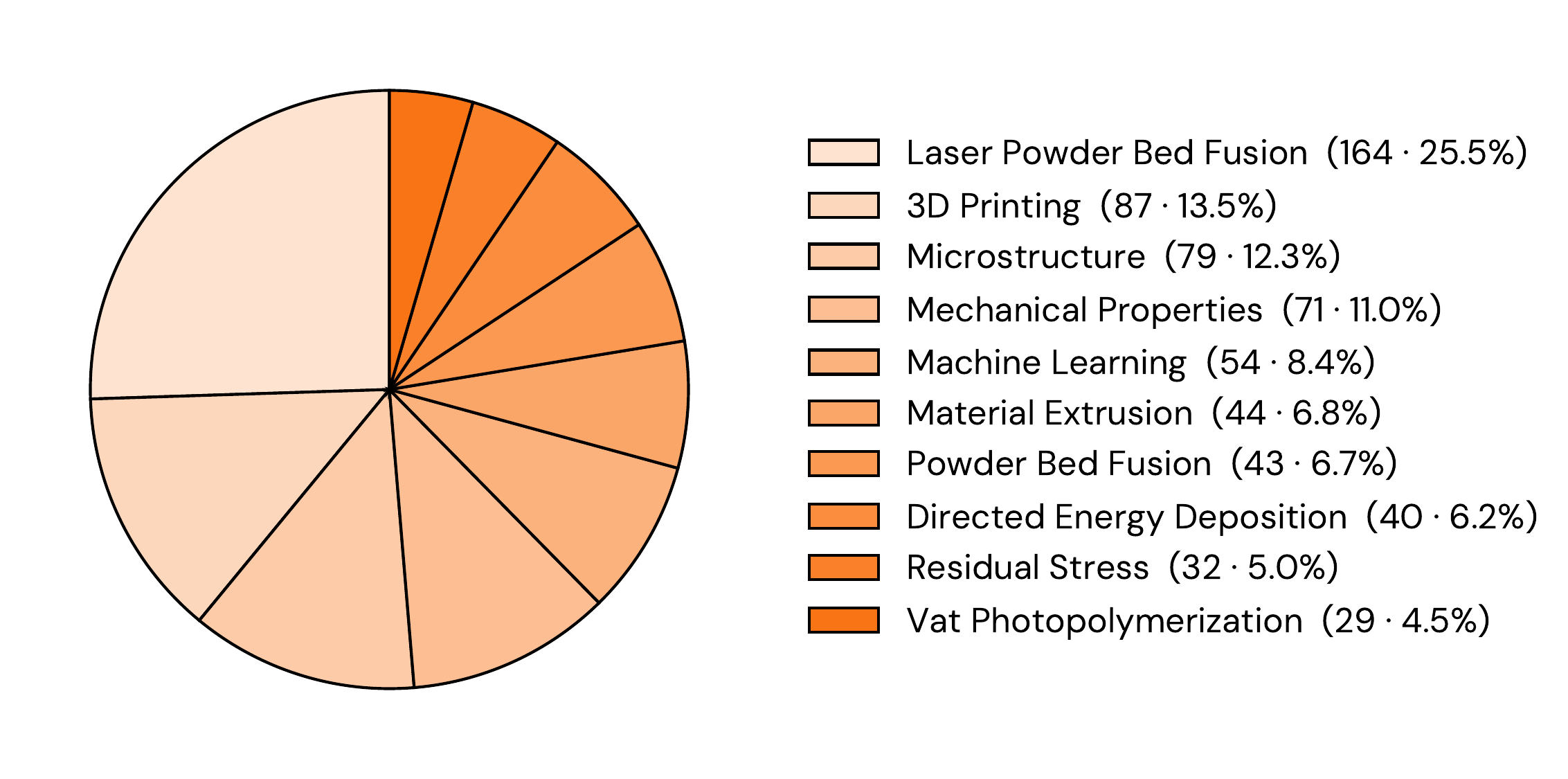}
    \caption{
        Plot of the top 10 keywords obtained from dataset articles omitting
        those where none are provided and ones where ``Additive Manufacturing"
        is listed.
    }
    \label{fig:keywords_top10}
\end{figure}

\section{Gemma 3 Pre-Trained Variant Results}
\label{sec:gemma_3_pt}
\begin{figure}
    \centering
    \includegraphics[width=\textwidth]{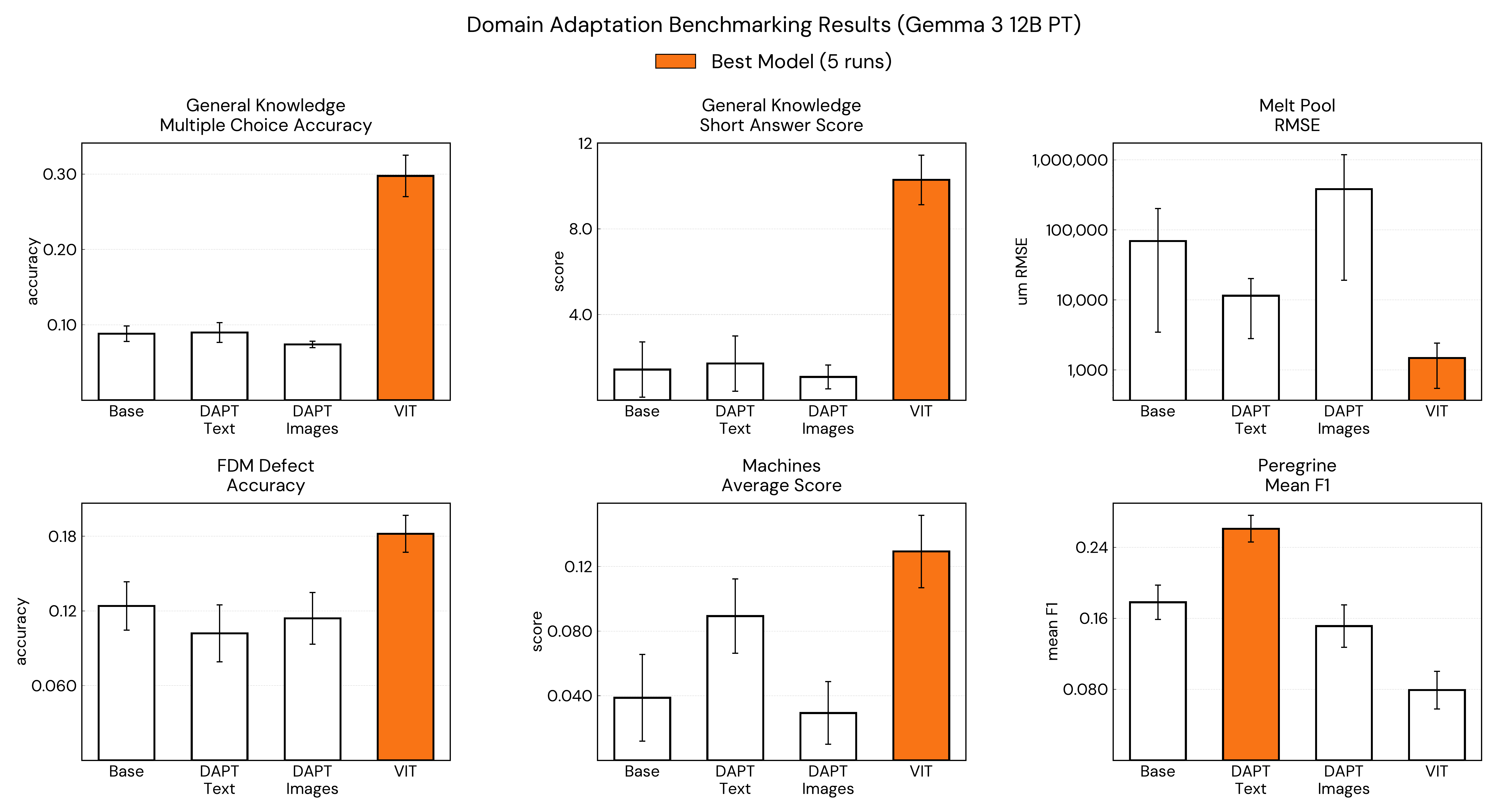}
    \caption{
        All models adapted upon the \texttt{Gemma-\allowbreak 3-\allowbreak
        12b-\allowbreak pt} (pretraining only) variant of the base model exhibit
        comparatively worse performance to their \texttt{Gemma-\allowbreak
        3-\allowbreak 12b-\allowbreak it} variants. Models at the visual
        instruction tuned achieve the best performance and in some cases
        outperform their previous training stages by a significant factor.
    }
    \label{fig:gemma_3_results_pt}
\end{figure}

With models developed from the base Gemma 3 model \cite{team_gemma_2025} with
only pretraining applied (Fig. \ref{fig:gemma_3_results_pt}), the performance on all
tasks were often significantly worse than those from the instruction tuned base.
For these cases, visual instruction tuned models showcased the best performance
in all tasks with the exception of LPBF anomaly identification where the
performance seems to decrease with additional training stages. This underlines
the impact that instruction tuning has on the general usability of an LLM and
the effect that visual instruction tuning can have to a specific domain.

One trend seen within all tasks for both the PT and IT variants of the base
model show that performance at the DAPT image training stage noticeably
decreases. This is seen most in language based tasks such as general knowledge
and in a few image based tasks such as LPBF anomaly identification. Contributing
factors to this could be the freezing of language attention weights during the
DAPT image stage, some images not having associated caption pairs, or
catastrophic forgetting.
\clearpage

\bibliography{references}

\end{document}